\documentclass[letterpaper]{article} 
\usepackage{aaai23}  
\usepackage{times}  
\usepackage{helvet}  
\usepackage{courier}  
\usepackage[hyphens]{url}  
\usepackage{graphicx} 
\usepackage{natbib}  
\usepackage{caption} 
\usepackage{algorithm}

\usepackage{newfloat}
\usepackage{listings}
\DeclareCaptionStyle{ruled}{labelfont=normalfont,labelsep=colon,strut=off} 
\floatstyle{ruled}
\newfloat{listing}{tb}{lst}{}
\floatname{listing}{Listing}
\usepackage{subfig}
\usepackage{tikz, pgf}      
\usepackage{algorithmicx} 
\usepackage{algpseudocode}
\usepackage{amsmath, bm, amssymb}

\usepackage{booktabs}
\usepackage{multirow}
\usepackage{makecell}
\usepackage{longtable}
\usepackage{supertabular}

\definecolor{TUMBlue}{HTML}{0065BD}
\definecolor{TUMSecondaryBlue}{HTML}{005293}
\definecolor{TUMSecondaryBlue2}{HTML}{003359}
\definecolor{TUMBlack}{HTML}{000000}
\definecolor{TUMWhite}{HTML}{FFFFFF}
\definecolor{TUMDarkGray}{HTML}{333333}
\definecolor{TUMGray}{HTML}{808080}
\definecolor{TUMLightGray}{HTML}{CCCCC6}
\definecolor{TUMAccentGray}{HTML}{DAD7CB}
\definecolor{TUMAccentOrange}{HTML}{E37222}
\definecolor{TUMAccentGreen}{HTML}{A2AD00}
\definecolor{TUMAccentLightBlue}{HTML}{98C6EA}
\definecolor{TUMAccentBlue}{HTML}{64A0C8}

\usepackage{pgfplots}
\usepackage{pgfplotstable}
\usepackage{tikz}
\tikzset{>=stealth}
\usetikzlibrary{patterns}
\usetikzlibrary{pgfplots.statistics}
\usetikzlibrary{positioning, shapes}
\usetikzlibrary{calc}

\tikzstyle{neuron} = [circle, minimum size=0.5cm, text width=0.2cm, text height=0.2cm, text centered, align=center, font=\fontsize{18}{0}\selectfont]
\tikzstyle{loss} = [very thick, rectangle, minimum width=3cm, minimum height=1.5cm, text centered, draw=black, align=center, font=\fontsize{18}{0}\selectfont]

\tikzstyle{module} = [rectangle, minimum width=2.5cm, minimum height=2cm, text centered, draw=black, rounded corners, align=center, font=\fontsize{16}{22}\selectfont]
\tikzstyle{transmodule} = [thick, rectangle, minimum width=2.5cm, minimum height=2cm, text centered, draw=black, rounded corners, align=center]
\tikzstyle{frame} = [very thick, rectangle, minimum width=2cm, minimum height=2cm, text centered, draw=black, rounded corners, align=center, font=\fontsize{16}{0}\selectfont, fill=TUMAccentBlue!10]
\tikzstyle{key} = [very thick, rectangle, minimum width=1.4cm, minimum height=1.4cm, text centered, rounded corners, align=center, font=\fontsize{16}{0}\selectfont, fill=TUMAccentGray!20]
\tikzstyle{dark_module} = [rectangle, minimum width=2.5cm, minimum height=2cm, text centered, draw=black, rounded corners, align=center, fill=TUMAccentLightBlue]

\tikzstyle{arrow} = [thick, ->,>=stealth,rounded corners=4pt, draw=black, align=center]
\tikzstyle{bluearrow} = [very thick,->,>=stealth,rounded corners=4pt, draw=TUMBlue, align=center]
\tikzstyle{darkgrayarrow} = [very thick,->,>=stealth,rounded corners=4pt, draw=TUMDarkGray, align=center]
\tikzstyle{dashedarrow} = [very thick,->,>=stealth,rounded corners=4pt, draw=black, align=center, dashed]
\tikzstyle{graydashedarrow} = [thick,->,>=stealth,rounded corners=4pt, draw=black, align=center, dashed, color=TUMGray]
\tikzstyle{orangedashedarrow} = [thick,->,>=stealth,rounded corners=4pt, draw=black, align=center, dashed, color=TUMAccentOrange]

\tikzstyle{marker} = [circle, minimum size=0.4cm, scale=0.8, text centered, draw=black, align=center, fill=orange!10, font=\large]

	\newcommand{\drawmain}{
		\centering  
		\begin{tikzpicture} [scale=0.18, transform shape]

			\node(all)[module, minimum width=48.5cm, minimum height=12cm, draw=black, fill=TUMBlue!10] at (-1.25,-0.5) {};
			\node(encoder)[module, dashed, minimum width=15cm, minimum height=11cm, draw=black, fill=TUMBlue!10] at (-16.5,-0.5) {};
			\path (encoder.north)+(-2,1.5) node (enc) {\fontsize{45}{0}\selectfont Task Inference Module};
			
			\node(gru)[module, thick, minimum width=7cm, minimum height=8cm, fill=TUMAccentLightBlue, font=\fontsize{45}{0}\selectfont] at (-20,-0.5) {Encoder \\[2ex]   };
			\path (-20, -2) node[scale=4.8] {$q_{\bm \phi}$};
			
			\node(buffer)[module, thick, minimum width=7cm, minimum height=3cm, rounded corners=0.6mm, font=\fontsize{45}{0}\selectfont] at (-31, -0.5) {Buffer};

			\node(y-pred)[module, thick, minimum width=2.4cm, minimum height=2.4cm, fill=TUMWhite] at (-11, 3) {};
			\path (-11, 3) node[scale=4] {$\bm y$};

			\node(mu-pred)[module, thick, minimum width=2.4cm, minimum height=2.4cm, fill=TUMWhite] at (-11, -0.5) {};
			\path (-11, -0.5) node[scale=4] {$\bm \mu$};
			
			\node(sigma-pred)[module, thick, minimum width=2.4cm, minimum height=2.4cm, fill=TUMWhite] at (-11, -4) {};
			\path (-11, -4) node[scale=4] {$\bm \sigma$};

			\node(gmvae)[module, thick, minimum width=7cm, minimum height=10.5cm, draw=TUMDarkGray, fill=TUMWhite] at (0,-0.5) {};
			\node(contrastive)[loss, thick, minimum width=8cm, minimum height=3.5cm, font=\fontsize{80}{80}\selectfont] at (0,8.5) {};
			\path (0, 8.5) node[scale=4.5] {$\mathcal{L}_{contrastive}$};

			\node(prob1) [loss, thin, minimum width=0.7cm, minimum height=1.2cm] at (-1.6, 1.5) {};
			\node(prob2) [loss, thin, minimum width=0.7cm, minimum height=1.0cm] at (-0.9, 1.4) {};
			\node(prob4) [loss, thin, minimum width=0.7cm, minimum height=1.4cm] at (1.9, 1.6) {};
			\node(prob3) [loss, thick, minimum width=0.7cm, minimum height=2.6cm, draw=TUMSecondaryBlue] at (1.2, 2.2) {};
			\path (0.2,1.6) node(y-dis) {\fontsize{45}{0}\selectfont ...};
			\draw[thick] (-2.8,0.9)--(2.8, 0.9);
			\path (0, 0.0) node[scale=4.5] {$q(\bm  y_{t})$};

			\def\normal-f{\x,{1.8/exp(2*(\x+0.1)^2)-3.4}}
			\draw[thick, draw=TUMSecondaryBlue, domain=-1.6:1.4] plot (\normal-f) node[right] {};
			
			\def\normal1{\x,{1/exp(2*(\x+1.5)^2)-3.6}}
			\draw[color=black, domain=-3:0] plot (\normal1) node[right] {};
			\def\normal2{\x,{1.2/exp(0.6*(\x)^2)-3.7}}
			\draw[color=black, domain=-2.2:2.2] plot (\normal2) node[right] {};
			\def\normal3{\x,{1.5/exp(3*(\x-1)^2)-3.6}}
			\draw[color=black, domain=0:2] plot (\normal3) node[right] {};
			
			\draw[thick] (-2.8,-3.7)--(2.8, -3.7);
			\path (0, -4.8) node[scale=4.5] {$q(\bm  z_{t})$};

			\node(sr-decoder)[module, thick, minimum width=7cm, minimum height=8cm, fill=TUMAccentLightBlue, font=\fontsize{45}{0}\selectfont] at (17,-0.5) {Decoder\\[2ex]    };
			\path (17, -2) node[scale=4.8] {$p_{\bm \phi}$};
			
			\node(loss-decoder)[loss, thick, minimum width=8cm, minimum height=3.5cm, font=\fontsize{52}{0}\selectfont] at (34,-0.5) {};
			\path (34, -0.5) node[scale=4.5] {$\mathcal{L}_{decoder}$};
			
			\path (9.5, 0.5) node[scale=5] {$\bm z_{t}$};

			\node(sac)[module, thick, minimum width=10cm, minimum height=6.5cm, fill=TUMGray!10, font=\fontsize{45}{0}\selectfont] at (17,10) {SAC \\[2ex]   };
			\path (17, 8.5) node[scale=4.8] {$\pi_{\bm \theta}$};
			
			\path (sac.north)+(0,1) node (po) {\fontsize{45}{0}\selectfont Policy Module};
			\node(loss-sac)[loss, thick, minimum width=8cm, minimum height=3.5cm, font=\fontsize{52}{0}\selectfont] at (34,10) {};
			\path (34, 10) node[scale=4.5] {$\mathcal{L}_{SAC}$};
			\path (9.5, 12) node[scale=5] {$\bm s_{t}$};
			
			\draw[arrow] (gru.east)--(mu-pred.west);
			\draw[arrow] (gru.east)--([xshift=1cm]gru.east)--([xshift=1cm]$(y-pred.west -|gru.east)$)--(y-pred.west);
			\draw[arrow] (gru.east)--([xshift=1cm]gru.east)--([xshift=1cm]$(sigma-pred.west -|gru.east)$)--(sigma-pred.west);
			
			\draw[arrow] ([yshift=0.3cm]encoder.east)--([yshift=0.3cm]gmvae.west);
			\draw[orangedashedarrow, <-] ([yshift=-0.3cm]encoder.east)--([yshift=-0.3cm]gmvae.west);
			
			\draw[arrow] ([xshift=-0.3cm]gmvae.north)--([xshift=-0.3cm]contrastive.south);
			\draw[orangedashedarrow, <-] ([xshift=0.3cm]gmvae.north)--([xshift=0.3cm]contrastive.south);
			
			\draw[arrow] (buffer.east)--(gru.west);
			\draw[arrow] (buffer.north)--([yshift=4cm]buffer.north)--([xshift=0cm, yshift=1cm]$(sac.west -|buffer.north)$)--([yshift=1cm]sac.west);
			
			\draw[arrow] (gmvae.east)--(sr-decoder.west);
			\draw[arrow] (gmvae.east)--([xshift=3cm]gmvae.east)--([xshift=3cm, yshift=-1cm]$(sac.west -|gmvae.east)$)--([yshift=-1cm]sac.west) node[midway,below, color=black, scale=5]{$q(\bm z_{t})$};
			
			\draw[orangedashedarrow, <-] ([yshift=-0.6cm]gmvae.east)--([yshift=-0.6cm]sr-decoder.west);
			
			\draw[arrow] ([yshift=0.3cm]sr-decoder.east)--([yshift=0.3cm]loss-decoder.west) node[midway,above,color=black, scale=5]{$\bm s_{t}, r_{t}$};
			\draw[orangedashedarrow, <-] ([yshift=-0.3cm]sr-decoder.east)--([yshift=-0.3cm]loss-decoder.west);
			
			\draw[arrow] ([yshift=0.3cm]sac.east)--([yshift=0.3cm]loss-sac.west);
			\draw[orangedashedarrow, <-] ([yshift=-0.3cm]sac.east)--([yshift=-0.3cm]loss-sac.west);

		\end{tikzpicture}
	}
	
	\newcommand{\drawdecoder}{
	\centering
	\begin{tikzpicture} [scale=0.32, transform shape]
		\node(decoder)[module, thick, minimum width=11cm, minimum height=6cm, fill=TUMBlue!10, label={[align=center, font=\fontsize{25}{0}\selectfont]MoSS-Decoder \\ [1ex]}] at (0.5,0) {};
		
		\node(state-dec)[module, thick, minimum width=5cm, minimum height=2cm, fill=TUMAccentLightBlue, font=\fontsize{22}{22}\selectfont] at (-1,1.3) {State \\[0.6ex] Decoder};
		\node(reward-dec)[module, thick, minimum width=5cm, minimum height=2cm, fill=TUMAccentLightBlue, font=\fontsize{22}{22}\selectfont] at (-1,-1.3) {Reward \\[0.6ex] Decoder};

		\path (-10, 0) node[scale=3](context) {($\bm s_{t}, \bm a_{t}, \bm z_{t}$)};
		\node(outs) [neuron, scale=1.7] at (3.5, 1.2) {$\hat{\bm s}_{t+1}$};
		\node(outr) [neuron, scale=1.7] at (4, -1.4) {$r_{t}$};
		
		\node(lossstate) [loss, minimum width=3.5cm, font=\fontsize{27}{27}\selectfont] at (9.5, 1.3) {};
		\path (9.5, 1.3) node[scale=2.3] {$\mathcal{L}_{state}$};
		\node(lossreward) [loss, minimum width=3.5cm, font=\fontsize{27}{27}\selectfont] at (9.5, -1.3) {};
		\path (9.5, -1.3) node[scale=2.3] {$\mathcal{L}_{reward}$};

		\node(true-s) [neuron, scale=1.7] at (14, 1.2) {$\hat{\bm s}_{t+1}$};
		\node(true-r) [neuron, scale=1.7] at (14.5, -1.4) {$r_{t}$};
		
		\draw[arrow] (context.east)--([xshift=1cm]context.east)--([xshift=1cm]$(state-dec.west -| context.east)$)--(state-dec.west);
		\draw[arrow] (context.east)--([xshift=1cm]context.east)--([xshift=1cm]$(reward-dec.west -| context.east)$)--(reward-dec.west);
		\draw[arrow] (state-dec.east) -- ($(outs.west)+(0, 0.1)$);
		\draw[arrow] (reward-dec.east) -- ($(outr.west)+(0, 0.1)$);
		\draw[arrow] ($(outs.east)+(1.2, 0.1)$) -- (lossstate.west);
		\draw[arrow] ($(outr.east)+(0.7, 0.1)$) -- (lossreward.west);
		\draw[arrow] ($(true-s.west)+(0, 0.1)$) -- (lossstate.east);
		\draw[arrow] ($(true-r.west)+(0, 0.1)$) -- (lossreward.east);
\end{tikzpicture}}

	\newcommand{\drawcontrastive}{
	\centering
	\begin{tikzpicture} [scale=0.35, transform shape]
		\node(query)[module, minimum width=2cm, minimum height=2.5cm, label={[align=center, font=\fontsize{22}{0}\selectfont]Query\\[.5ex]}] at (2,0) {};
		\path (2, 0) node[scale=2.5] {$\bm Q$};
		
		\node(keys)[module, minimum width=9cm, minimum height=2.5cm, fill=TUMBlue!10, label={[align=center, font=\fontsize{22}{0}\selectfont]Keys\\[.5ex]}] at (9,0) {};

		\node(key-1)[module, thick, minimum width=1.7cm, minimum height=1.7cm] at (6, 0) {};
		\path (6, 0) node[scale=2.5] {$\bm k_1$};
		\node(key-2)[module, thick, minimum width=1.7cm, minimum height=1.7cm] at (8.5, 0) {};
		\path (8.5, 0) node[scale=2.5] {$\bm k_2$};
		\path (10.5, 0) node[scale=3.5] {$\cdots$};   
		\node(key-n)[module, thick, minimum width=1.7cm, minimum height=1.7cm] at (12, 0) {};
		\path (12, 0) node[scale=2.5] {$\bm k_{N_k}$};   
		
		\node(pos)[module, minimum width=4.5cm, minimum height=2.7cm, draw=TUMBlue, fill=TUMBlue!30, dashed, font=\fontsize{25}{0}\selectfont] at (4.5,-5) {Positive \\[1ex] Pairs};
		\node(neg)[module, minimum width=4.5cm, minimum height=2.7cm, draw=TUMDarkGray, fill=TUMDarkGray!20, dashed, font=\fontsize{25}{0}\selectfont] at (10,-5) {Negative \\[1ex] Pairs};
		
		\node(loss)[module, minimum width=8cm, minimum height=2.3cm, font=\fontsize{25}{0}\selectfont] at (7.25,-10) {Contrastive Loss};
		
		\draw[arrow, color=TUMBlue] (query.south) -- (pos.north);
		\draw[arrow, color=TUMBlue] (key-1.south) -- (pos.north);
		
		\draw[arrow] (query.south) -- (neg.north);
		\draw[arrow] (key-2.south) -- (neg.north);
		\draw[arrow] (key-n.south) -- (neg.north);
		
		\draw[arrow, color=TUMBlue] (pos.south) -- (loss.north);
		\draw[arrow] (neg.south) -- (loss.north);
	\end{tikzpicture}
}

\title{Meta-Reinforcement Learning Based on Self-Supervised Task Representation Learning}
\author{
    Mingyang Wang,\textsuperscript{\rm 1} Zhenshan Bing,\textsuperscript{\rm 1} Xiangtong Yao,\textsuperscript{\rm 1} Shuai Wang,\textsuperscript{\rm 2} \\
    Hang Su,\textsuperscript{\rm 3} Chenguang Yang,\textsuperscript{\rm 4}, Kai Huang,\textsuperscript{\rm 5, 6} \thanks{Corresponding Author} Alois Knoll\textsuperscript{\rm 1} 
}

\affiliations{
    \textsuperscript{\rm 1}Department of Informatics, Technical University Munich, \textsuperscript{\rm 2}Tencent Robotics X Lab,\\ \textsuperscript{\rm 3}Dipartimento di Elettronica, Politecnico di Milano, \textsuperscript{\rm 4}Bristol Robotics Laboratory, University of the West of England\\ \textsuperscript{\rm 5}School of Computer Science and Engineering, Sun Yat-Sen University, \textsuperscript{\rm 6}Shenzhen Institute, Sun Yat-Sen University\\

    mingyang.wang@tum.de, bing@in.tum.de, xiangtong.yao@tum.de, shuaiwanghit@gmail.com, \\ hang.su@polimi.it, cyang@ieee.org, huangk36@mail.sysu.edu.cn, knoll@in.tum.de
}

\usepackage{bibentry}

\begin{document}

\maketitle

\begin{abstract}
Meta-reinforcement learning enables artificial agents to learn from related training tasks and adapt to new tasks efficiently with minimal interaction data. However, most existing research is still limited to narrow task distributions that are parametric and stationary, and does not consider out-of-distribution tasks during the evaluation, thus, restricting its application. In this paper, we propose MoSS, a context-based \textbf{M}eta-reinforcement learning algorithm based \textbf{o}n \textbf{S}elf-\textbf{S}upervised task representation learning to address this challenge.We extend meta-RL to broad non-parametric task distributions which have never been explored before, and also achieve state-of-the-art results in non-stationary and out-of-distribution tasks. Specifically, MoSS consists of a task inference module and a policy module. We utilize the Gaussian mixture model for task representation to imitate the parametric and non-parametric task variations. Additionally, our online adaptation strategy enables the agent to react at the first sight of a task change, thus being applicable in non-stationary tasks. MoSS also exhibits strong generalization robustness in out-of-distributions tasks which benefits from the reliable and robust task representation. The policy is built on top of an off-policy RL algorithm and the entire network is trained completely off-policy to ensure high sample efficiency. On MuJoCo and Meta-World benchmarks, MoSS outperforms prior works in terms of asymptotic performance, sample efficiency (3-50x faster), adaptation efficiency, and generalization robustness on broad and diverse task distributions.
\end{abstract}
\section{Introduction}
Modern deep reinforcement learning (RL) has made significant progress in learning complex behavior \cite{Mnih2015, AlphaGO, AlphaZERO, HGG}. However, they typically do not transfer learned skills to other tasks, require to re-train the policy for new tasks. In contrast, humans can learn new skills efficiently using prior knowledge and experience. Motivated by this, meta-reinforcement learning (meta-RL) was developed to mimic the human learning process by learning a prior model from a set of related training tasks and quickly adapt to unseen tasks during testing.


However, most existing meta-RL studies are severely confined to narrow task distributions that are parametric and stationary, let alone taking into consideration out-of-distribution (OOD) test tasks. In prior works \cite{MAML, E-MAML, Reptile, ProMP, RL2, MAESN, PEARL, MQL, asTaskInference, Focal,  VariBAD, Hyper-X}, task distributions only involve parametric variations, e.g., simulated robots can adapt to reach a new goal velocity after being trained on a set of different goal velocities previously. It is therefore unreasonable to expect generalization to a completely new control task, e.g., reaching a specified goal position. To extend the meta-RL adaptation to qualitatively distinct tasks, training tasks should also include such non-parametric variability in which task differences cannot be expressed only using continuous parameters. Also, humans can quickly adapt their behavior to unexpected changes and perturbations, we expect the meta-RL algorithm to be broadly useful in non-stationary scenarios where the task may vary at any time step and exhibit strong generalization robustness for out-of-distribution tasks. 

As a promising approach to tackling the meta-RL problem, context-based meta-RL \cite{RL2, PEARL, MQL, asTaskInference, VariBAD, MCAT} extracts salient information from past experience and generates latent representations on which the policy is then conditioned. However, representing tasks on broad task distributions poses new challenges to the model as more complex relationships and dependencies between tasks must be captured. Existing studies \cite{PEARL, VariBAD, MQL, Focal} often uses a single-component Gaussian for task representation which is inadequate in such complex cases.

Furthermore, current meta-RL algorithms suffer from sample inefficiency due to on-policy optimization \cite{MAML, E-MAML, Reptile, ProMP, RL2, MAESN, VariBAD, Hyper-X}, or adaptation inefficiency due to the few-shot adaptation strategy \cite{PEARL, MQL, Focal}. Although on-policy RL approaches are easy to incorporate into meta-RL, they require a large quantity of data for training, making them sample inefficient. Recent works \cite{PEARL, MQL, Focal} integrated the off-policy RL algorithm into the meta-RL framework, resulting in a significant increase in sample efficiency. However, they still suffer from adaptation inefficiency as task representation is updated at the trajectory level, thus enable to accomplish non-stationary tasks.

With these considerations in mind, we propose MoSS, a context-based meta-RL algorithm based on self-supervised task representation learning. The method aims to be (1)  applicable to diverse task distributions with parametric and non-parametric, stationary and non-stationary, in-distribution, and out-of-distribution tasks. The general performance of meta-RL algorithms on such broad task distributions has never been explored before, and (2) sample efficient by using as few data samples as possible for training, and adaptation efficient by taking as few environmental steps as possible to adapt to a new task during meta-testing.

MoSS is made up of two modules: a task inference module that takes the task trajectory as input and encodes it as latent variables, and a policy module that uses the task representation to explore the environment and learn optimal actions. First, in the task inference module, we realize representative and robust task inference using the Gaussian mixture latent space for task representation and the contrastive learning strategy. By extending the VAE-based inference network with a Gaussian mixture latent space, MoSS accommodates non-parametric variations using different Gaussian clusters and parametric variations with variability within each Gaussian component. Additionally, contrastive learning enhances the ability of MoSS to differentiate different tasks while clustering similar ones. Reliable task representation is a prerequisite for effective downstream RL. Second, we use the gated recurrent unit (GRU) \cite{GRU} as the encoder and optimize it with the next-step reconstruction loss to achieve online adaptation. Unlike prior works \cite{PEARL, VariBAD} that generate task variables based on trajectory-level context, our encoder updates latent variables based on local context, so that the agent can quickly adapt to new tasks and solve non-stationary tasks. Third, we build our policy on top of the soft-actor-critic(SAC) algorithm and condition the policy on the agent state and the latent task variable, to account for task uncertainty in its decision-making. We use a shared data buffer for task inference and policy training to realize a fully off-policy optimization that ensures high sample efficiency. We evaluate MoSS on MuJoCo \cite{Mujoco} and Meta-World \cite{MetaWorld} benchmarks, including various robotic control and manipulation tasks. MoSS shows state-of-the-art results in asymptotic performance, sample and adaptation efficiency, and generalization robustness\footnote{Implementation and videos available at \url{https://sites.google.com/view/metarl-moss}}.

\section{Background}

\subsection{From RL to Meta-RL}
In standard RL, a task is formulated as a Markov Decision Process (MDP) \cite{MDP} denoted as $M = (\mathcal{S}, \mathcal{A}, P, R, \gamma)$ where $\mathcal{S}$ is a state space, $\mathcal{A}$ is an action space, $P(\bm s^{'}|\bm s, \bm a)$ denoting the transition function, $R(r|\bm s, \bm a)$ the reward function and $\gamma$ the discount factor. The objective is to find an optimal policy $\pi$ that maximizes the expected return in this MDP. Original RL algorithms are mostly trained on one task at a time. While meta-RL leverages a set of related tasks to learn a policy that can quickly adapt to unseen test tasks. Specifically, during meta-training, the algorithm has access to $N_{train}$ tasks drawn from the task distribution $p(M)$. At meta-test time, new tasks are also sampled from $p(M)$. The meta-trained policy should quickly adapt to new tasks to achieve maximum return, i.e., $\bm \theta_{*} = \arg \max_{\bm \theta} \mathbb{E}_{M \sim p(M)}\Big[\mathbb{E}_{\bm \tau \sim p(\bm \tau|\pi_{\bm \theta})}\big[\sum_{t \geq 0} \gamma^{t}r_{t}\big]\Big]$. Here $\bm \tau_{:t} = \{\bm s_{t}, \bm a_{t}, r_{t}, \bm s_{t+1}\}_{0:t}$ represents the task trajectory that collects the agent's transitions $(\bm s_{t}, \bm a_{t}, r_{t}, \bm s_{t+1})$ up to the current time step.
\begin{figure*}[t!]
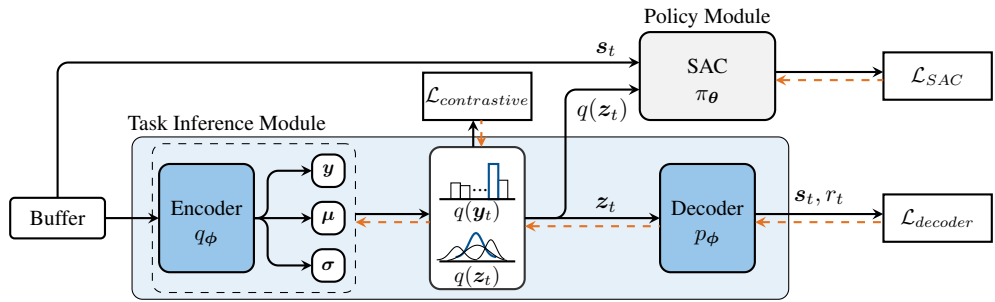

    \centering
    \drawmain
    \caption{MoSS Architecture: MoSS consists of a task inference module and a policy module. The inference module encodes the task context $\bm \tau$ as latent task distributions $q_{\bm \phi}(\bm z)$, then the policy module conditions on the agent state $\bm s$ and the task representation $q_{\bm \phi}(\bm z)$ to act in the environment.}
    \label{fig:moss-architecture}
\end{figure*}
\subsection{Meta-RL Environments}

\subsubsection{Parametric and non-parametric variability}The variability of task distributions is a key property in the meta-RL context. The formal definition of two properties of the task distribution is given in \cite{MetaWorld} based on the kind of structure tasks have in common. \textbf{Parametric variability} describes tasks that qualitatively share the same property, but the task parameterization varies, while \textbf{non-parametric variability} describes tasks that are qualitatively distinct. For example, in the \textit{Cheetah-Multi-Task} benchmark, the \textit{Cheetah-Velocity} task aims to control the robot to run at different goal velocities. All tasks essentially share the same qualitative task descriptions and are parameterized by the goal velocity. While \textit{Cheetah-Velocity} and \textit{Cheetah-Goal} tasks are qualitatively distinct. Although they both control the robot to exhibit certain behaviors, the difference between them cannot be described by parametric variations. Thus, they belong to different task families (or base tasks).


\subsubsection{Adaptation in meta-RL environments}When being exposed to new tasks, the meta-RL agent is usually allowed to collect context data for a few episodes to get an increasingly better task understanding. We use the term $k$-shot to represent the number of exploration episodes. The agent should identify tasks and adapt the policy within $k$ episodes, so $k$ is an important metric to evaluate the adaptation efficiency of meta-RL algorithms. Note that we use $k=0$ for MoSS, i.e., reported results are from the first evaluation rollout without any previous data collection, which is a harder experiment setting but also more realistic as the agent should solve the task at first sight without any failed trials. 

\section{Related Work}
Recent studies in meta-RL can be separated into two categories: optimization-based and context-based meta-RL. Optimization-based meta-RL \cite{MAML, E-MAML, Reptile, ProMP} learns a prior model based on training tasks and then performs task-specific fine-tuning on unseen tasks to quickly adapt to new tasks with a few gradient descent steps. Although the idea is conceptually elegant, the gradient descent update is uninformative for RL algorithms, making it hard to explore a new, unknown environment. In contrast, context-based meta-RL \cite{RL2, PEARL, MQL, asTaskInference, VariBAD, MCAT, David, bing2021meta, Alex, Xiangtong} adapts to new tasks by aggregating past experience into latent representations on which the policy is then conditioned. Such methods realize a better exploration-exploitation trade-off, as they maintain belief over possible MDPs that enable reasonable exploration by acting optimally according to these MDPs. Even without any explicit fine-tuning, the meta-RL model can capture some task characteristics and adapt to new test tasks.

As a context-based meta-RL method, RL$^{2}$ \cite{RL2} first proposed to condition the policy on hidden states encoded from task trajectory. By structuring the agent as a recurrent neural network, RL$^{2}$ enables online policy learning based on RNN dynamics. Another popular training mechanism is to disentangle the task inference and action selection process to enable explicit task representations. In PEARL \cite{PEARL}, task context is encoded as probabilistic latent variables via variational inference. Then using posterior sampling, latent task encodings are integrated with RL policy and guide the agent to act in the environment. In the following work \cite{asTaskInference}, the task inference network is trained in a supervised way and the policy directly conditions the posterior distributions to reason for task uncertainty. Closely related to our approach is VariBAD \cite{VariBAD}, which adopts VAE for task inference. The encoder processes the task context and generates latent task distributions, based on which the decoder predicts past and future states and rewards. The network is trained based on the prediction loss rather than using RL-loss \cite{PEARL} or using privileged task information\cite{asTaskInference}. Inspired by VariBAD, MoSS also uses VAE for task inference. However, it uses a different encoder-decoder training strategy. MoSS only reconstructs the next-step state and reward instead of modeling all past and future steps, i.e., we do not assume the task within one episode is always consistent. Therefore, MoSS can adjust its task belief to potential task changes within a single episode and is applicable to non-stationary environments. 

All the studies mentioned above only focus on the performance of meta-RL algorithms on commonly used parametric task distributions and do not explore the applicability of these algorithms on broad task distributions with non-parametric task variability, non-stationary environments, and out-of-distribution evaluation tasks. Some other works discuss the meta-RL performance in non-stationary or OOD tasks \cite{learning-to-adapt, MQL, OOD-1, LDM}. However, non-parametric task distributions have never been explored specifically, and no studies discuss these scenarios together to investigate the algorithms’ general performance. For the first time, MoSS explores the meta-RL algorithm applicable to broad and diverse task distributions.

\section{Methodology}
In the following, we first give an overview of our MoSS algorithm. Then we explain the strategy of MoSS in the task inference module and policy module to make it applicable to the aforementioned task distributions with good sample and adaptation efficiency. We also summarize the meta-training procedure of MoSS as pseudo-code in Algorithm \ref{alg:meta-training}.

\subsection{Algorithm Overview}
\label{sec:Algorithm Overview}
MoSS disentangles task inference from policy control as in \cite{PEARL, VariBAD, MELD}. The task inference module $q_{\bm \phi}(\bm z_{t}| \bm \tau_{:t}))$, parameterized by $\bm \phi$, encodes the task trajectory $\bm \tau_{:t}$ as latent task distributions $q_{\bm \phi}(\bm z_{t}))$, and it is trained using a decoder $p_{\bm \phi}(\bm s_{t+1}, r_{t}))$ that predicts next-step states and rewards from current states and actions. The policy module $\pi_{\theta}(a_{t}|s_{t}, q_{\bm \phi}(\bm z_{t}))$, parameteried by $\bm \theta$, conditions on the agent state $\bm s_{t}$ with the task belief $q_{\bm \phi}(\bm z_{t})$ to act in the environment.

The model architecture of MoSS is shown in Figure \ref{fig:moss-architecture}. To accommodate both the parametric and non-parametric task variations in the task inference module, we extend the VAE architecture with Gaussian mixture latent space for task representation. To further improve the model's ability to capture complex relationships between tasks, inspired by \cite{MCAT, Contrastive-2},we introduce contrastive loss as an auxiliary objective and jointly train the task inference network in a self-supervised manner. Moreover, to enable online adaptation, we use GRU \cite{GRU} as our encoder and update task representations at the transition level. In the policy module, following the Bayes-Adaptive MDP setting \cite{BAMDP, BayesianRL, VariBAD}, we condition the policy on the agent state augmented with the inferred task distribution to incorporate task uncertainty in its action selection process. Finally, we use a shared replay buffer for both modules and train the network in a fully off-policy manner.

MoSS works as follows: during meta-training, we sample trajectory data from training tasks and feed them into the task inference module, which learns to infer latent variables $q_{\bm \phi}(\bm z_{t})$ via self-supervised representation learning (details see Section \ref{sec:task-inference}). Also, the policy network $\pi_{\theta}(\bm a_{t}|\bm s_{t}, q_{\bm \phi}(\bm z_{t}))$ takes in the agent state $\bm s_{t}$ with the current task belief $q_{\bm \phi}(\bm z_{t})$ and learns to select optimal actions. While meta-testing, the encoder generates and updates task representations $q_{\bm \phi}(\bm z_{t}|\bm \tau_{:t})$ at each time step based on the collected task experience $\bm \tau_{:t}$. Then the policy $\pi_{\theta}(\bm a_{t}|\bm s_{t}, q_{\bm \phi}(\bm z_{t}))$ conditions its actions on the agent state $\bm s_{t}$ augmented with the task belief $q_{\bm \phi}(\bm z_{t})$ to interact with the environment.


\renewcommand{\algorithmicrequire}{\textbf{Input:}}
\begin{algorithm}[t!]
  \caption{MoSS Meta-training}\label{alg:algorithm1}
  \begin{algorithmic}[1]
    \renewcommand{\algorithmicrequire}{\textbf{Input:}}
    \Require Task distribution $p(M)$,\ encoder $q_{\bm\phi}$,\ decoder $p_{\bm\phi}$,\ actor $\pi_{\bm \theta}$,\ critic $Q_{\bm \omega}$,\ replay buffer $\mathcal{B}$
    \While{not done}
    \State Sample tasks $\bm M=\{M_{i}\}_{i=1}^{N}$ from $p(M)$
    \State Collect trajectories with $\pi_{\theta}$ and add to buffer $\mathcal{B}$ \\ \Comment{\textit{Data collection}}
    
    \For{step in training steps}\Comment{\textit{Training step}}
        \State Sample training tasks $\bm M_{train}$ from $p(M)$
        
        \For{$M_{i} \in \bm M_{train}$}
            \State Sample $\bm \tau_{:T} \sim \mathcal{B}_{i}$ with trajectory length $T$
            \State Infer task beliefs $\{ q_{\bm \phi}(\bm z_{t}|\bm \tau_{:t})\}_{0:T-1}$
            \State Calculate $ELBO_{i} = \sum_{t=0}^{T-1} ELBO_{i,\ t}(\bm \phi)$ using Eq. \ref{con:ELBO}
        \EndFor
        \State Calculate contrastive loss $\mathcal{J}_{cont}$ using Eq. \ref{con:infoNCE}
        \State Update $\bm \phi \gets \bm \phi - \alpha_{\bm \phi}\nabla_{\bm \phi}(\sum_{i}ELBO_{i} + \mathcal{J}_{cont})$\\ \Comment{\textit{Inference network update}}
        \State Update $(\bm \theta, \bm \omega)$ with SAC algorithm \\ \Comment{\textit{SAC update}}
    \EndFor
    \EndWhile
  \end{algorithmic}
  \label{alg:meta-training}
\end{algorithm}

\begin{figure*}[t!]
  \centering
    \includegraphics[width=1\textwidth]{{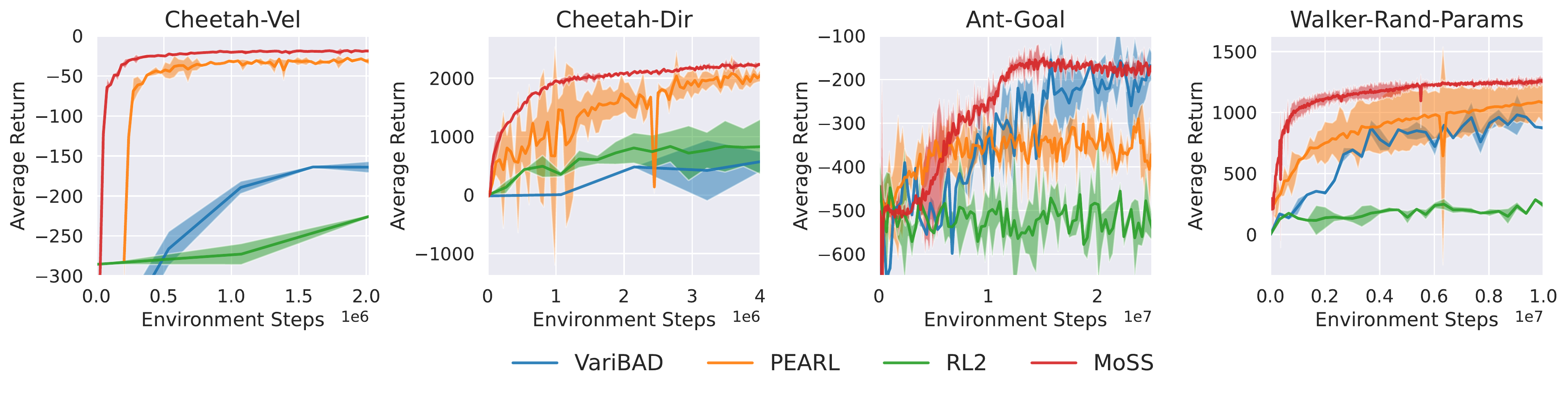}}
  \caption{Meta-test performance in parametric MuJoCo environments: Average return (y-axis) against collected environment steps during meta-training (x-axis).}
	\label{fig:mujoco-parametric-result}
\end{figure*}
\subsection{Task Inference Module}
\label{sec:task-inference}
\subsubsection{VAE with Gaussian Mixture Latent Space}
\label{sec:gaussian-mixture}

Previous context-based meta-RL works \cite{PEARL, MELD, VariBAD, asTaskInference} use single-component Gaussian distributions for task representation. However, representing tasks from complex task distributions poses new challenges to the inference network and single-component Gaussian-based representation is therefore insufficient. To address this problem, we construct a Gaussian mixture latent space that accommodates non-parametric variability with different Gaussian clusters and parametric variability using variations within each Gaussian component. We use a categorical variable $\bm y \sim Cat(\pi)$ to indicate the base task probability and latent variables $\bm z \sim N(\mu(\bm y), \sigma(\bm y))$ as base task-specific Gaussian components. Together with the input $\bm x$, the joint probability is factorized as $p(\bm x,\bm y,\bm z) = p(\bm y)p(\bm z|\bm y)p(\bm x|\bm z)$ \cite{VaDE, CURL-Rao}. The posterior inference of $p(\bm y, \bm z|\bm x)$ is intractable, we employ the learned approximate posterior $q(\bm y, \bm z|\bm x)=q(\bm z|\bm x, \bm y)q(\bm y| \bm x)$. The posterior inference and data generation process run as follows: first, given the input $\bm x$, the cluster inference model $q(\bm y|\bm x)$ produces the categorical distribution $q(\bm y)$, and task encoding models $q(\bm z|\bm x, \bm y^{(k)})$ generate K independent Gaussians for each cluster. Then, the decoder $p(\bm x|\bm z)$ reconstructs data $\bm x$ from the latent variable $\bm z$. Following the variational inference approach and Monte-Carlo approximation, the evidence lower bound objective (ELBO) is formulated as (derivation see Appendix):

\begin{equation}
\label{con:ELBO}
\begin{aligned}
ELBO \approx  &\sum_{k=1}^{K} \overbrace{q(\bm y^{(k)}|\bm x)}^{\text{\shortstack{Component \\ posterior}}} \Bigl[\overbrace{\log p(\bm x|\bm z^{(k)})}^{\text{\shortstack{Component-wise \\ reconstruction loss}}} \\
&-\alpha\overbrace{\mathbb{KL}(q(\bm z^{(k)}|\bm x, \bm y^{(k)})\| p(\bm z|\bm y^{(k)}))}^{\text{Component-wise regularizer}}\Bigr] \\
&-\beta\overbrace{\mathbb{KL}(q(\bm y|\bm x)\|p(\bm y))}^{\text{Categorical regularizer}}
\end{aligned}
\end{equation}

Here $\alpha$ and $\beta$ represent the regularization weight for the KL divergence term of the component-wise Gaussian distribution and the categorical distribution, respectively. Intuitively, the model can either have high entropy over $q(\bm y|\bm x)$, where all component-wise losses should be low, or assign a high $q(\bm y^{(k)}|\bm x)$ for some $k$ and use that component to model the data well. Here we introduce hyperparameters $\alpha$ and $\beta$ to weight the component-wise and categorical regularization terms. We parameterize the decoder as two independent networks: state decoder $p_{\bm \phi}(\bm s_{t+1}|\bm s_{t}, \bm a_{t}, \bm z_{t})$ and reward decoder $p_{\bm \phi}(\bm r_{t}|\bm s_{t}, \bm a_{t}, \bm z_{t})$. Both networks are modeled as regression networks, using the MSE loss between predictions ($\hat{\bm s}_{t+1}$, $\hat{\bm r}_{t}$) and true targets ($\bm s_{t+1}$, $r_{t}$) from buffer as supervision, therefore the networks are trained in a self-supervised manner. Unlike prior works \cite{RL2, PEARL}, we do not back-propagate the RL loss into the encoder, which completely decouples task inference from policy learning. This modular structure facilitates the encoder to identify tasks only based on task characteristics. The decoder is not used at meta-test time. During meta-test, given the trajectory up to the current time step $\bm \tau_{:t}$, the encoder predicts the $p(\bm y)$ distribution and Gaussian parameters $\{\bm \mu_{k}, \bm \sigma_{k}\}_{k=1\cdots K}$. We pick the Gaussian component $q(\bm z_{t}) = N(\bm \mu_{\bm z}(\bm y_{t}(k^{*})), \bm \sigma_{\bm z}(\bm y_{t}(k^{*})))$\ corresponding to the most likely base task $k^{*}=\arg \max_{k} \{q(\bm y_{t}^{(k)} |\bm \tau_{:t})\}_{k=1\cdots K}$ and get the best matching Gaussian at each time step.

\subsubsection{Contrastive Representation Learning}
\label{sec:contrastive-learning}

Essentially, we expect the encoder to cluster similar tasks and distinguish different tasks by organizing their latent embeddings in a structured way. This coincides with the idea of contrastive learning, which learns a latent embedding space where similar sample pairs stay close while dissimilar ones are far apart. Motivated by \cite{MCAT, Contrastive-2}, we introduce a contrastive learning objective for task inference training to realize better task identification. In contrastive learning, input data are usually organized into positive and negative pairs. We use the key-query definition in \cite{InstanceDiscrimination,CURL-contrastive,TCL}: a key-query pair is positive if they are representations belong to the same data instance and negative otherwise. Given a query $\bm Q$ and a set of keys $ \mathbb{K} = \{\bm k_{i}\}_{i=0…N_{K}}$ consisting of 1 positive key and $N_{K}-1$ negative keys, contrastive learning aims to ensure that the query $\bm Q$ matches with the positive key $\bm k_{+}$ more than any of the negative keys.

We use task embeddings $\bm z$ as queries and keys. Embeddings from the same task (but different time steps) build positive pairs, while embeddings from different tasks are negative pairs. Specifically, we sample two latent embeddings of different time steps from each task as the query and its corresponding positive key, and sample $N_{K}-1$ embeddings from other tasks as negative keys. We use the InfoNCE score \cite{InfoNCE} with the 
Euclidean distance as the similarity metric to calculate the contrastive objective:

\begin{equation}
\label{con:infoNCE}
\mathcal{J}_{cont} = \frac{1}{N_Q}\sum_{i=0}^{N_Q-1}\log \frac{\exp(\text{sim}(\bm Q_{i}, \bm k_{+}))}{\begin{matrix} \sum_{j=0}^
{N_{K}-1}\end{matrix} \exp(\text{sim}(\bm Q_{i}, \bm k_{j}))}
\end{equation}

Here $N_Q$ is the number of sampled queries, and $N_{K}$ is the number of keys for each query. The objective can be interpreted as the log-likelihood of an $N_{K}$-way softmax classifier where $\bm k_{+}$ is the label of the corresponding query. Then we get the final objective $\mathcal{J}(\bm \phi) = {ELBO} + \mathcal{J}_{cont}$.

\subsubsection{RNN-based Online Inference}
Moreover, we introduce a recurrent version of the VAE \cite{VRNN} to map the sequential input to time step-wise latent variables. Specifically, we model a VAE at each time step and explicitly study the dependencies between latent variables across consecutive time steps. Instead of sharing a global prior distribution $p_{0}(\bm y) $ and $p_{0}(\bm z) $ for $\bm y$ and $\bm z$ of all time steps (as in \cite{PEARL}), we use posterior distributions from the previous time step $q(\bm z|\bm x_{<t}, \bm y)$ and $q(\bm y|\bm x_{<t})$ as current prior. The resulting time step-wise variational lower bound ELBO is given in Appendix.
%

We use the GRU \cite{GRU} as the encoder $q_{\bm \phi}(\bm z_{t}|\bm \tau_{:t})$. By recursively reasoning with its hidden states $q_{\phi}(\bm z_{t}|\bm \tau_{:t}) = q_{\phi}(\bm z_{t}|\bm s_{t}, \bm a_{t}, \bm r_{t}, \bm h_{t-1})$, MoSS combines historical information with the current input and adjusts actions step by step. On the one hand, it enables more efficient task inference and policy adaptation. On the other hand, MoSS is, therefore, applicable to non-stationary environments where the task can potentially change at any time step. 
\begin{figure*}[t]
  \centering
    \subfloat[Training Curve] {\includegraphics[width=0.485\textwidth]{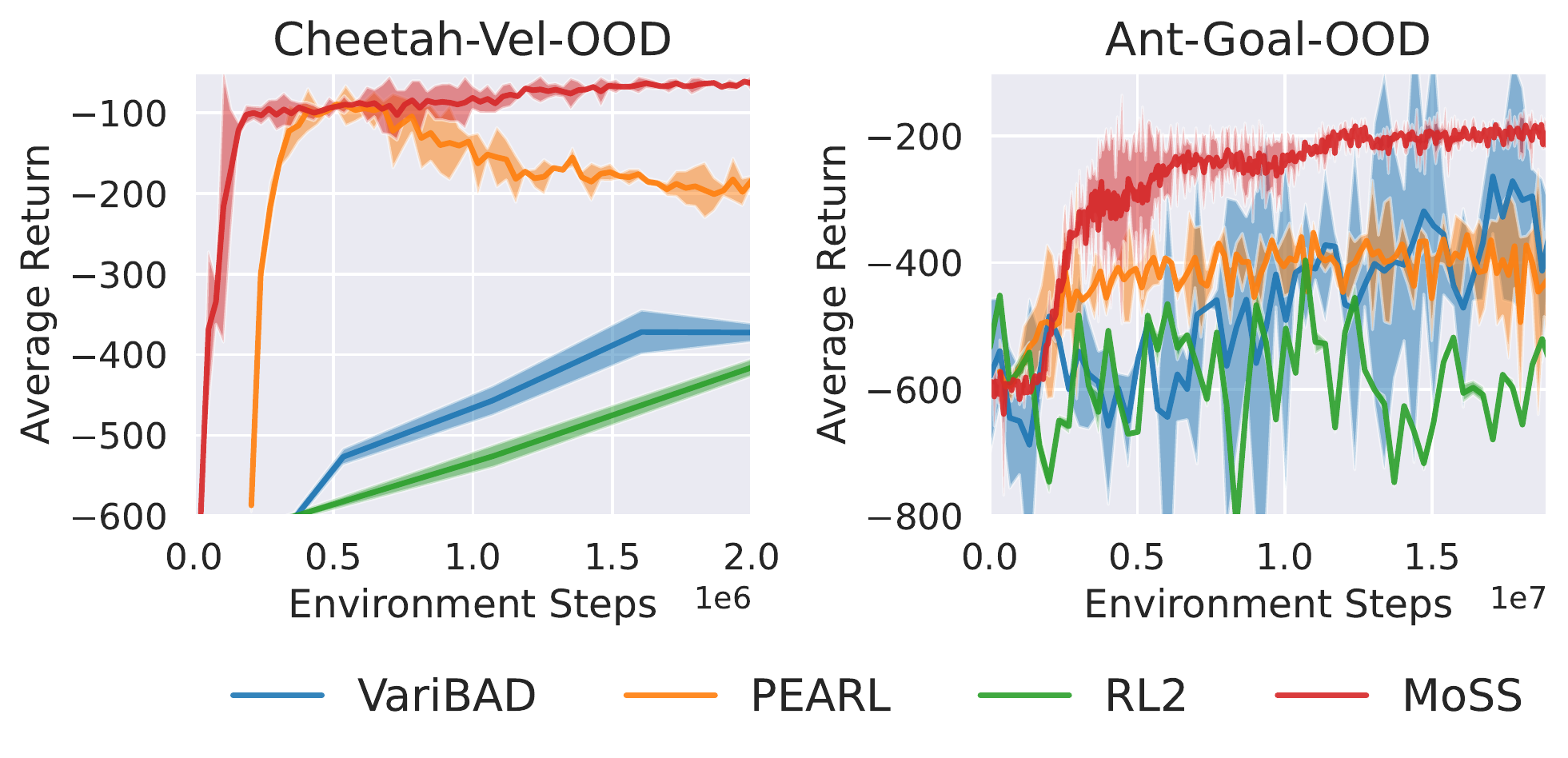}}
    \subfloat[PEARL embeddings] {\includegraphics[width=0.25\textwidth]{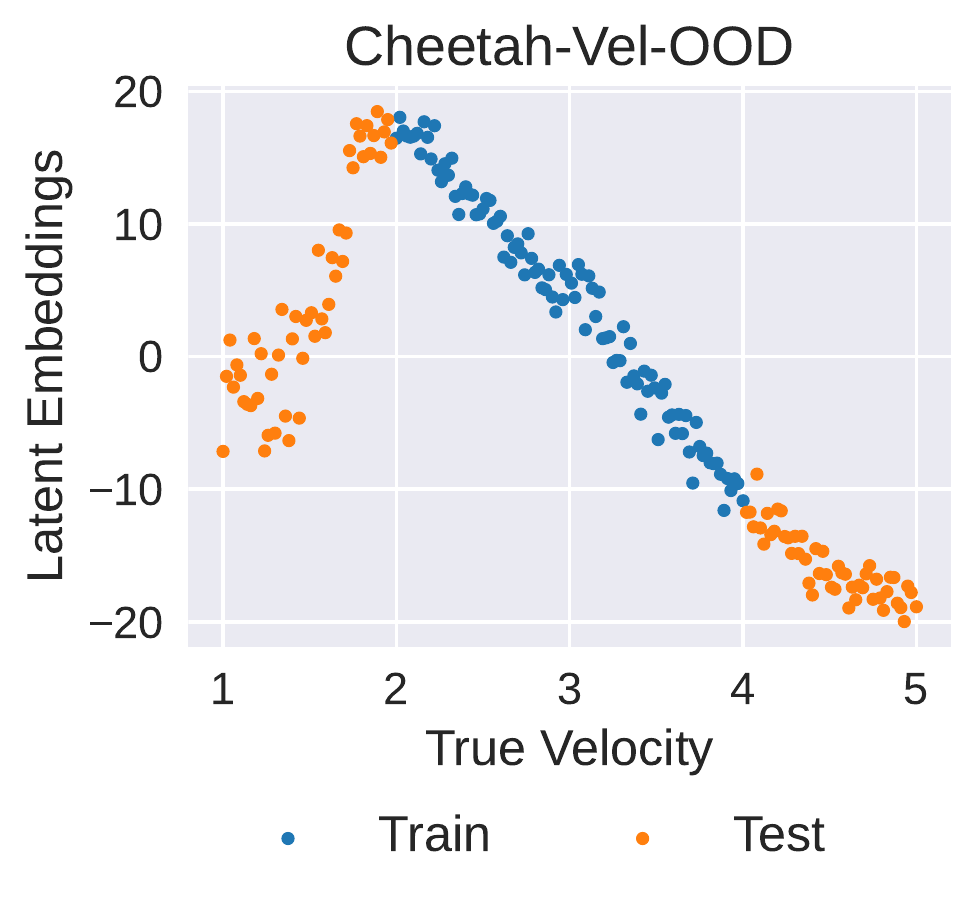}}
    \subfloat[MoSS embeddings] {\includegraphics[width=0.25\textwidth]{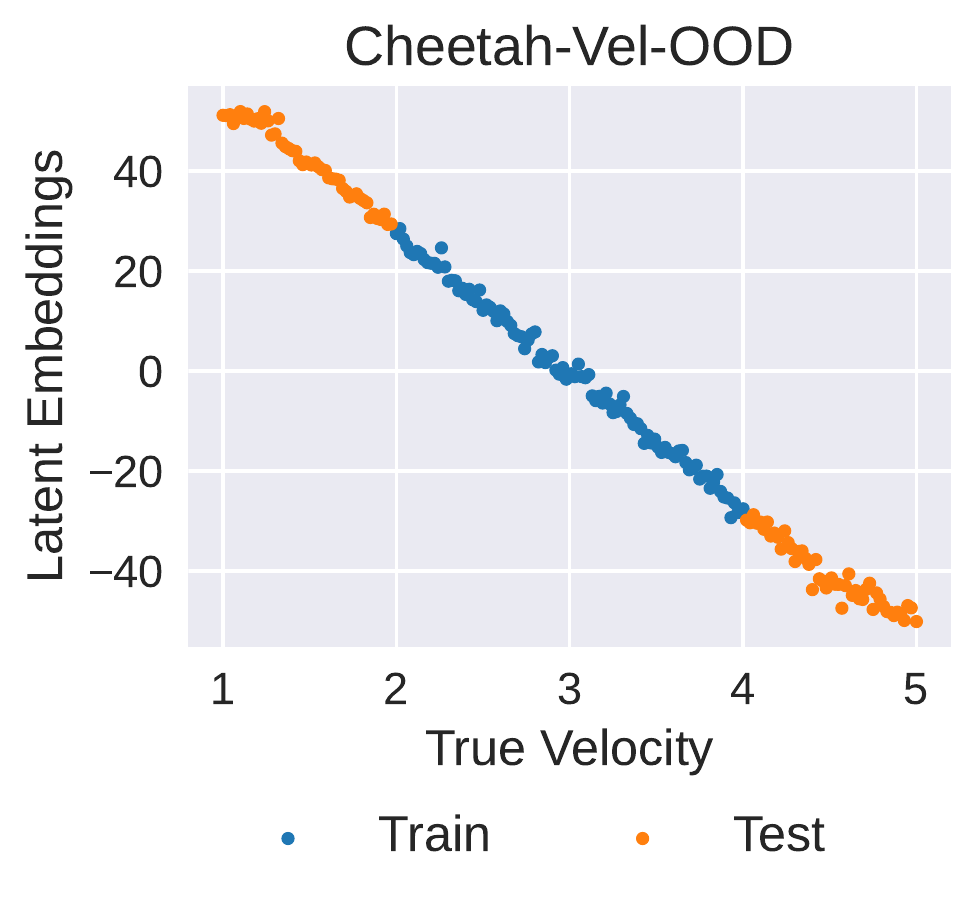}}

{\caption{Meta-test performance in parametric MuJoCo environments with OOD tasks}\label{fig:ood-result}}
\end{figure*}

\begin{figure*}[htbp]
  \centering
    \includegraphics[width=0.716\textwidth]{{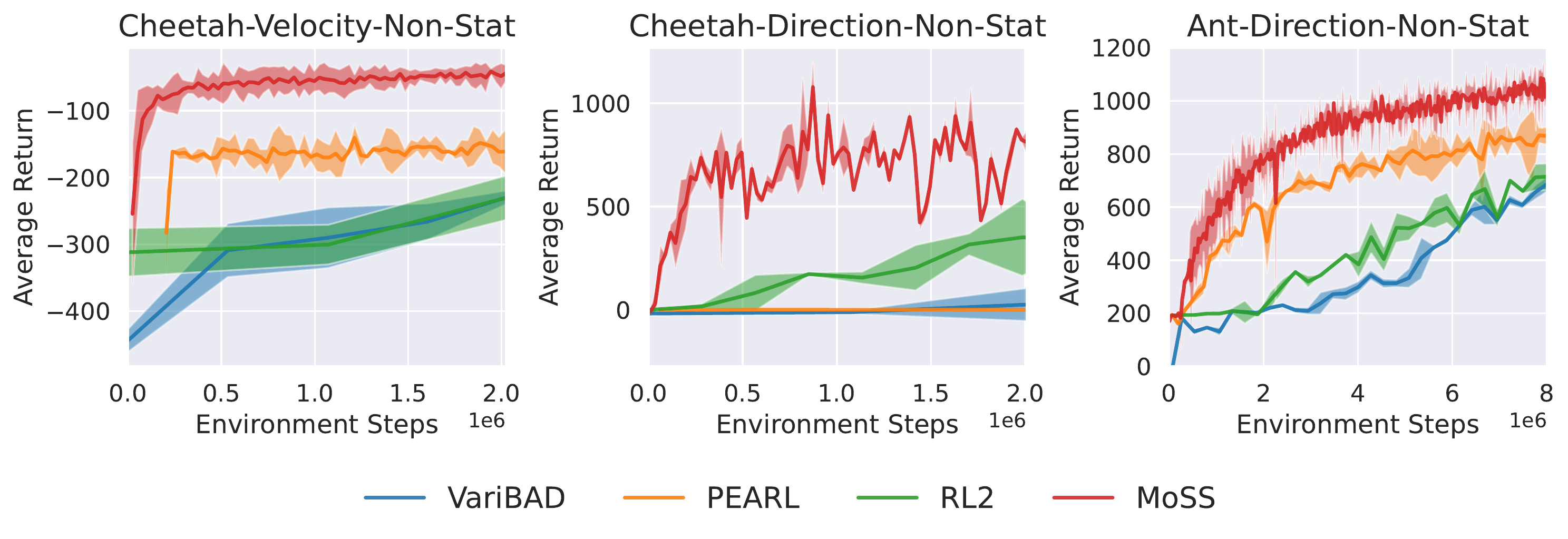}}
    \includegraphics[width=0.26\textwidth]{{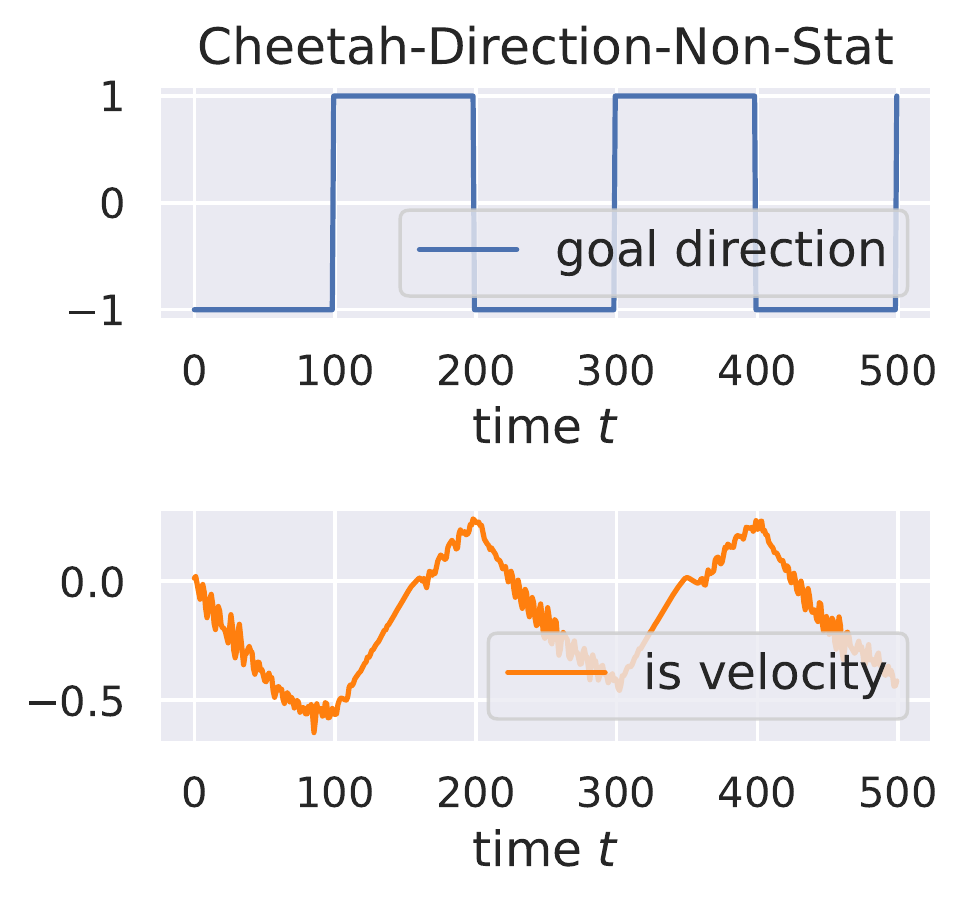}}
  \caption{Meta-test performance in non-stationary MuJoCo environments: (Left) Training curves; (Right) Agent Response in the \textit{Cheetah-Direction-Non-Stat} environment.}
	\label{fig:non-stationary-result}
\end{figure*}

\subsection{Policy Module}
We adopt the same strategy with \cite{VariBAD} in the policy module by interpreting the meta-RL problem as a planning problem in Bayes-Adaptive MDPs (BAMDPs) \cite{BAMDP}. A BAMDP is described by $(\mathcal{S}^{+}, \mathcal{A}, P, R, \gamma)$ where the hyper-state $\bm s_{t}^{+} \in \mathcal{S}^{+} = \mathcal{S} \times \mathcal{B}$ consists of the agent state $\bm s_{t}$ and the task belief $\bm b$. In BAMDPs, the RL policy conditions not only on the agent state but also on the agent's belief about the environment. A Bayes-optimal agent learns to maximize the expected return by systematically seeking out actions to reduce its environmental uncertainty (\textit{exploration}) and then taking promising actions to maximize the expected return (\textit{exploitation}) \cite{VariBAD}. Thus, the policy automatically learns how to trade off exploration and exploitation under task uncertainty.

In MoSS, the hyper-state is denoted as $\bm s_{t}^{+} = (\bm s_{t}, q_{\bm \phi}(\bm z_{t}))$ with $q_{\bm \phi}(\bm z_{t})$ from the task inference module. It directly incorporates the task uncertainty, unlike the policy $\pi_{\bm \theta}(\bm a|\bm s, \bm z)$ in \cite{MAESN, PEARL, MELD} uses $\bm z \sim q_{\bm \phi}(\bm z)$ via posterior sampling. We build our policy on top of the SAC \cite{SAC-1} and use a shared data buffer for VAE and RL training to realize fully off-policy training, unlike \cite{PEARL} that still uses an on-policy VAE training buffer, which makes MoSS outperform prior algorithms with respect to sample efficiency.

\section{Experiments}
\label{sec:experiments}
We evaluate the performance of MoSS on Mujoco \cite{Mujoco} and Meta-World \cite{MetaWorld} benchmarks. We compare MoSS with RL$^{2}$, PEARL and VariBAD \cite{RL2, PEARL, VariBAD}. First, we run experiments on commonly used parametric task distributions. Additionally, to verify the broad applicability of MoSS, we run experiments on diverse task distributions with out-of-distribution test tasks, non-stationary environments and non-parametric variations. We also evaluate several design choices of MoSS through ablation in Appendix. Note that we conduct the zero-shot meta-test on MoSS ($k=0$). In contrast, for MuJoCo tasks, VariBAD and RL$^{2}$ use $k=2$ and PEARL uses $k=3$. For Meta-World tasks numbers are given in Table \ref{tab:meta-world-result}. All the choices of $k$ are the same as their original papers.  We truncate the $x-$axis at the number of time steps required for MoSS to converge; for the full timescale results see Appendix. Other hyperparameters can be found in Appendix.

\begin{table}[htbp]
  \centering
    \begin{tabular}{c|c|c|c|c}
    \toprule
    Method & $k$-th episode & Reach & Push  & Pick-Place \\
    \midrule
    RL$^{2}$   & 10    & \textbf{100} & 96 & 98 \\
    PEARL & 10    & 68    & 44    & 28 \\
    \textbf{MoSS} & \textbf{1}     & 86    & \textbf{100}    & \textbf{100} \\
    \bottomrule
    \end{tabular}%
  \caption{Meta-World V2 ML1 result comparison: Success rate results are given in percentage\footnotemark}
  \label{tab:meta-world-result}
\end{table}


\begin{figure*}[t!]
  \centering
    \includegraphics[width=0.36\textwidth]{{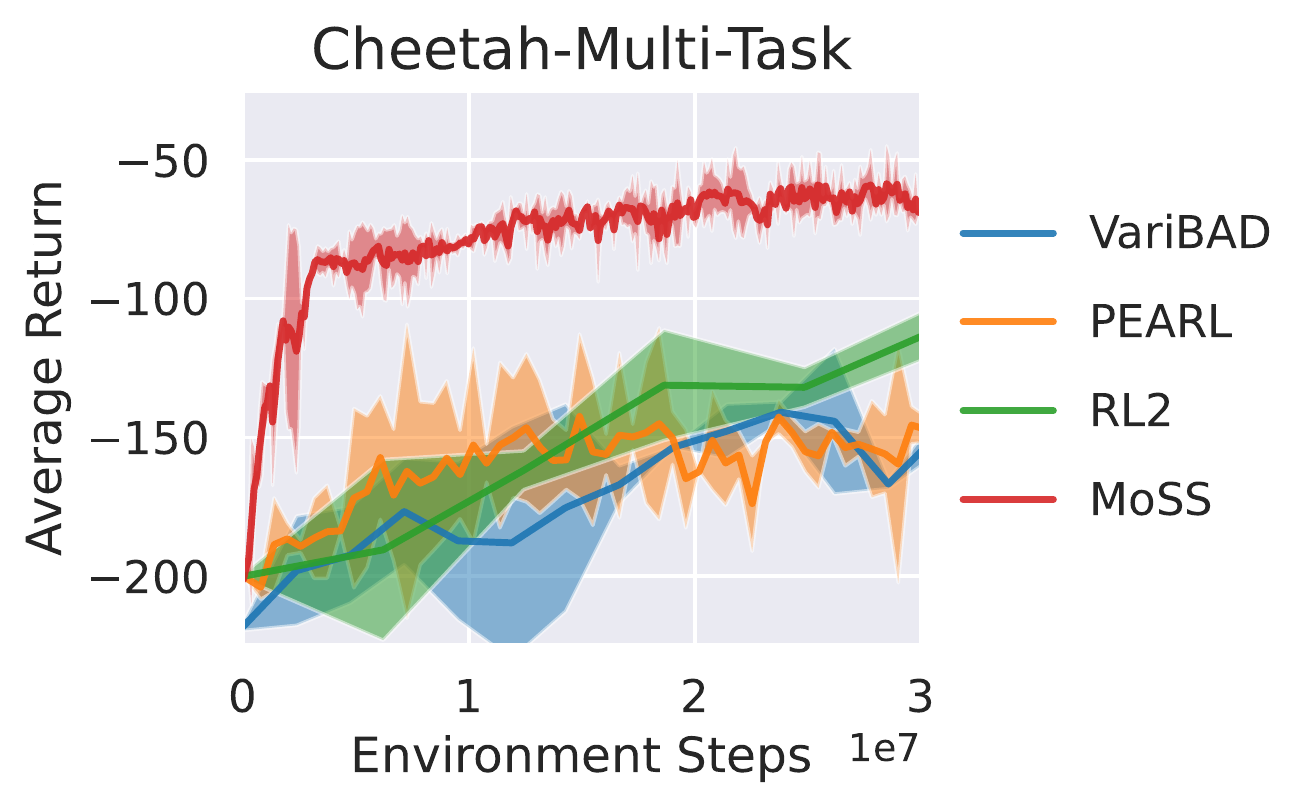}}
    \includegraphics[width=0.36\textwidth]{{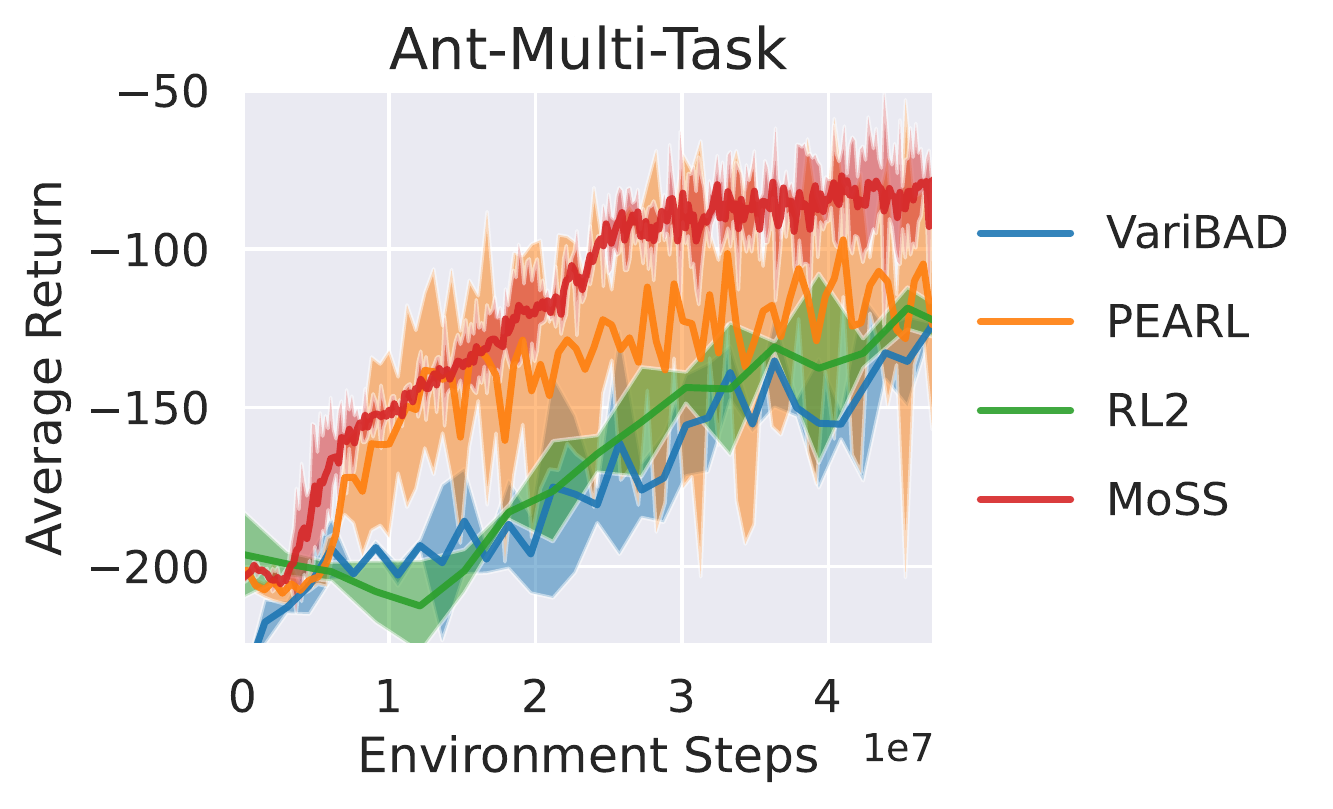}}
    \includegraphics[width=0.36\textwidth]{{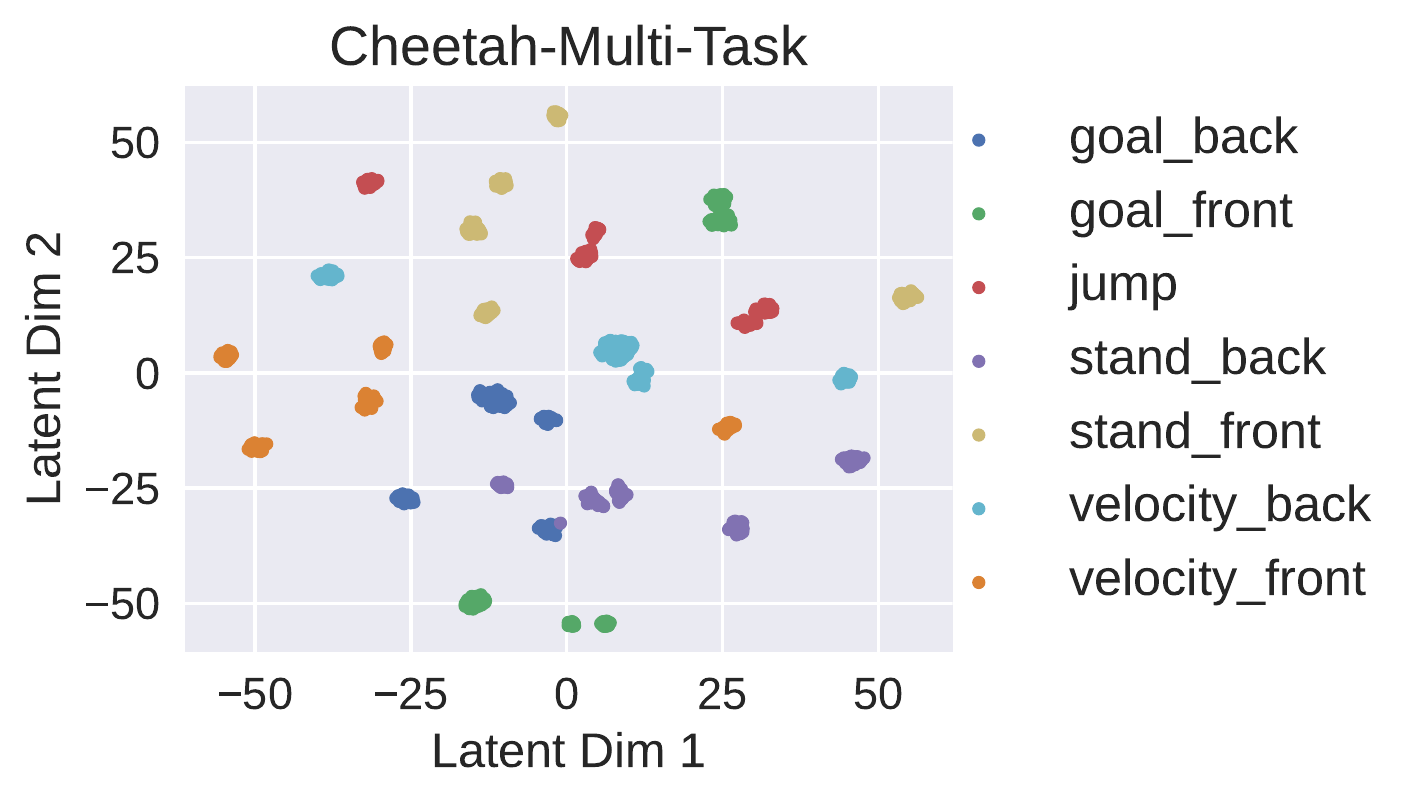}}
    \includegraphics[width=0.36\textwidth]{{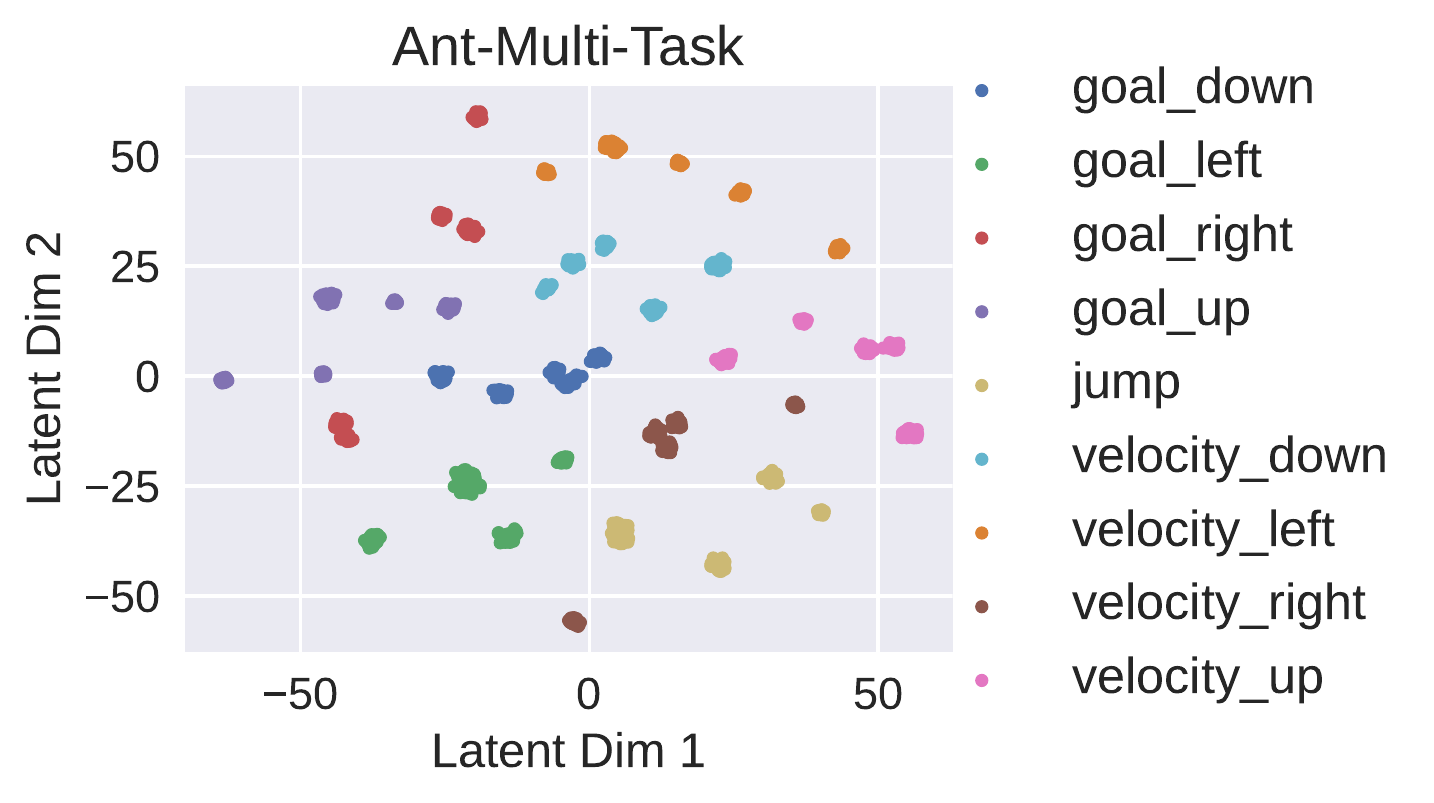}}
  \caption[Meta-test performance in non-parametric MuJoCo environments]{Meta-test performance in non-parametric MuJoCo environments}
	\label{fig:non-parametric-result}
\end{figure*}

\label{sec:parametric-result}
\subsubsection{Parametric task distributions} 
To validate the performance of MoSS in parametric environments, we first compare MoSS with baseline methods on MuJoCo and Meta-World ML1 tasks. Results are given in Figure \ref{fig:mujoco-parametric-result} and Table \ref{tab:meta-world-result}. MuJoCo locomotion tasks require the model to adapt across various reward functions or system dynamics (task details see Appendix). Meta-World ML1 benchmark consists of three robotic manipulation tasks where task variations are specified goal positions and use success rate as the evaluation metric. Training curves are given in Appendix. 



 \footnotetext{Baseline results are taken from \cite{MetaWorld} as the performance is highly sensitive to hyperparameters and hard to reproduce. VariBAD \cite{VariBAD} only uses the original Meta-World V1, for a fair comparison we also include results comparison on Meta-World V1 in Appendix.}


\subsubsection{Parametric Task Distributions with Out-of-Distribution Test Tasks} To further investigate the generalization robustness of our algorithm, we set up two OOD environments where we distinguish task ranges during meta-training and meta-test. For example, in \emph{Cheetah-Vel-OOD}, we train the agent on the velocity range of $[2.0, 4.0]$ and test it on $[1.0, 2.0] \cup [4.0, 5.0]$.  For the detailed setup see Appendix.

As plotted in Figure \ref{fig:ood-result} (left), MoSS outperforms baseline methods as it achieves higher average returns and better stability. We suppose this comes from the robust task representation of MoSS. Faced with new OOD test tasks, MoSS can still capture the task characteristics, which share similarity with previously seen tasks, and produce understandable representations for the downstream policy network. In Figure \ref{fig:ood-result} (right), we visualize the latent space of PEARL and MoSS on \emph{Cheetah-Vel-OOD} to demonstrate our assumption. The $x-$axis is the ground-truth target velocity and the $y-$axis is the t-SNE \cite{t-sne} visualization of latent embeddings (1D). Over the entire velocity range, MoSS shows a clear linear relationship between the latent embeddings and true targets. While PEARL struggles to figure out how to represent OOD test tasks, especially in the velocity range of $[1.0, 2.0]$, which makes it hard for the RL policy to learn optimal actions under unclear instruction.



\label{sec:non-stationary results}
\subsubsection{Parametric, Non-stationary Task Distributions} We also set up three non-stationary environments where the goal velocity or direction changes randomly every 100 time step. We choose the changing steps of 100 as the agent needs some time steps to adjust its behavior. For this reason, we increase the time horizon to 500~600 in non-stationary tasks. Additionally, at every 100 steps we hold a change probability $p=0.5$, i.e., the task changes at every 100 time step at the probability of 0.5. Randomness is necessary for non-stationary tasks otherwise the agent would remember the changing patterns and cannot adapt well to dynamic changes in more general non-stationary environments. Given the changing randomness, the task is different in each episode so that the agent will not remember the fixed changing pattern. In Figure \ref{fig:non-stationary-result}, we plot the experiment results and visualize the agent response to task changes in \emph{Cheetah-Direction-Non-Stat}, more visualizations see Appendix.



\subsubsection{Non-parametric Task Distributions} Finally, we set up two non-parametric benchmarks \emph{Cheetah-Multi-Task} and \emph{Ant-Multi-Task}. Both of them consist of multiple base tasks that are qualitatively distinct and sub-tasks inside each base task. The detailed task description is given in Appendix.

As Figure \ref{fig:non-parametric-result} shows, MoSS outperforms prior works in both environments. We use t-SNE \cite{t-sne} to visualize the latent embeddings of MoSS in two dimensions in Figure \ref{fig:non-parametric-result}. We use different colors to represent sub-tasks with different goal directions from the same base task as they usually exhibit dissimilar task characteristics, e.g., \textit{goal\_back} and \textit{goal\_front} are both in base task \textit{Cheetah-Goal}, but we plot them in different colors.  Each refined base task consists of 5 sub-tasks with different goal values corresponding to small clusters in the picture. MoSS can differentiate different tasks and cluster embeddings from the same task. Given explicit task representations as guidance, it is easier for the policy to learn optimal actions.




\section{Conclusion}
In this paper, we propose MoSS, an algorithm to address the meta-RL problem on diverse task distributions with superior sample and adaptation efficiency. We showed that reliable and robust task representations are crucial for meta-RL, especially on complex task distributions. Our approach represents tasks as a mixture of Gaussians in the latent space to accommodate parametric and non-parametric task variations. Together with contrastive learning, we realize effective task identification, which alleviates the difficulty of downstream policy learning. Furthermore, MoSS can rapidly adapt to new tasks in non-stationary environments via GRU-based online task inference and adaptation. We train the network completely off-policy to ensure high sample efficiency. MoSS outperforms existing methods in terms of asymptotic performance, sample and adaptation efficiency, and generalization robustness on various robot control and manipulation tasks with the single episode result. 

\section{Acknowledgements}
This work was supported by the Shenzhen Basic Research Grants (JCYJ20180508152434975, JCYJ20180507182508857), the European Union's Horizon 2020 Framework Programme for Research and Innovation under the Specific Grant Agreement No. 945539 (Human Brain Prohect SGA3).

\bibliography{aaai23}
\clearpage

\appendix

\section{Appendix}
\subsection{Full ELBO Derivation}
\label{sec:elbo-derivation}

Equation \ref{con:ELBO} can be derived as follows.
\begin{equation}
\begin{aligned}
\log p(\bm x_{:t}) &=\ \log \int_{\bm y, \bm z} p(\bm x_{:t}, \bm y, \bm z)\ dy dz \\
&=  \log \int_{\bm y, \bm z} p(\bm x_{:t}, \bm y, \bm z) \frac{q(\bm y, \bm z|\bm x_{:t})}{q(\bm y, \bm z|\bm x_{:t})}\ dy dz \\
&=\ \log \mathbb{E}_{q(\bm y, \bm z|\bm x_{:t})} \biggl[\frac{p(\bm x_{:t}, \bm y, \bm z)}{q(\bm y, \bm z|\bm x_{:t})}\biggr] \\
&\geq\ \mathbb{E}_{q(\bm y, \bm z|\bm x_{:t})} \Bigl[\log p(\bm x_{:t}, \bm y, \bm z) - \log q(\bm y, \bm z|\bm x_{:t})\Bigr] \\
&=\ \mathbb{E}_{q(\bm z|\bm x_{:t}, \bm y) q(\bm y|\bm x_{:t})} \Bigl[\log p(\bm x_{:t}|\bm z) + \log p(\bm z|\bm y) \\ &\ \ \ \ \ + \log p(\bm y) - \log q(\bm z|\bm x_{:t}, \bm y) - \log q(\bm y|\bm x_{:t})\Bigr] \\
&=\ \mathbb{E}_{q(\bm z|\bm x_{:t}, \bm y) q(\bm y|\bm x_{:t})} \Bigl[\log p(\bm x_{:t}|\bm z)\Bigr] \\ & \ \ \ \ \ -\mathbb{E}_{q(\bm z|\bm x_{:t}, \bm y) q(\bm y|\bm x_{:t})} \biggl[\frac{q(\bm z|\bm x_{:t}, \bm y)}{p(\bm z|\bm y)}\biggr]  \\ & \ \ \ \ \  -\mathbb{E}_{q(\bm y|\bm x_{:t})} \biggl[\frac{q(\bm y|\bm x_{:t})}{p(\bm y)}\biggr] \\
&=\ \mathbb{E}_{q(\bm y|\bm x_{:t})} \biggl[ \mathbb{E}_{q(\bm z|\bm x_{:t}, \bm y)} \Bigl[\log p(\bm x_{:t}|\bm z)  \\ & \ \ \ \ \  -\mathbb{KL}(q(\bm z|\bm x_{:t}, \bm y)\| p(\bm z|\bm y))\Bigr] - \mathbb{KL}(q(\bm y|\bm x_{:t})\|p(\bm y))\biggr] \\
&=\ ELBO_{t}
\end{aligned}
\end{equation}  

Using Monte-Carlo sampling, the ELBO can then be formulated as:

\begin{equation}
\begin{aligned}
ELBO \ \approx \  &\sum_{k=1}^{K} q(\bm y^{(k)}|\bm x) \Bigl[\log p(\bm x|\bm z^{(k)}) \\ & -\alpha\ \mathbb{KL}(q(\bm z^{(k)}|\bm x, \bm y^{(k)})\| p(\bm z|\bm y^{(k)}))\Bigr] \\
&-\beta\ \mathbb{KL}(q(\bm y|\bm x)\|p(\bm y))
\end{aligned}
\end{equation}

Further, we adopt a recurrent version of VAE \cite{VRNN} to explicitly restrict the divergence between posterior variables across adjacent time steps. Specifically, we use the posterior probability from the previous time step $q(\bm z|\bm x_{:t-1}, \bm y)$ as the current prior. Then we get the ELBO:

\begin{equation}
\label{con:ELBO-GMVAE-SEQ}
\begin{aligned}
ELBO_{t}\ \approx\  &\sum_{k=1}^{K} q(\bm y_{i}(k)|\bm x_{i, :t}) \Bigl[\log p(\bm x_{i, :t}|\bm z_{i}^{(k)})  \\ &-\alpha\ \mathbb{KL}(q(\bm z_{i}|\bm x_{i, :t}, \bm y_{i}(k))\|p(\bm z_{i}|\bm y_{i}(k),  \bm x_{i, <t}))\Bigr] \\
&-\beta\ \mathbb{KL}(q(\bm y_{i}|\bm x_{i,:t})\|p(\bm y_{i}|\bm x_{i,<t}))
\end{aligned}
\end{equation}

\subsection{Additional Results}

\subsubsection{Ablation Study}
\label{sec:ablation-study}

Finally, we ablate several design choices of our approach to better understand the essential features of MoSS. The ablation experiments are conducted on \emph{Cheetah-Vel-OOD} and \emph{Cheetah-Multi-Task} environments, as they are representative task distributions with parametric and non-parametric variability, respectively.
\label{fig:online-inference-ablation}
\begin{figure*}[t!]
  \centering
    \includegraphics[width=0.8\textwidth]{{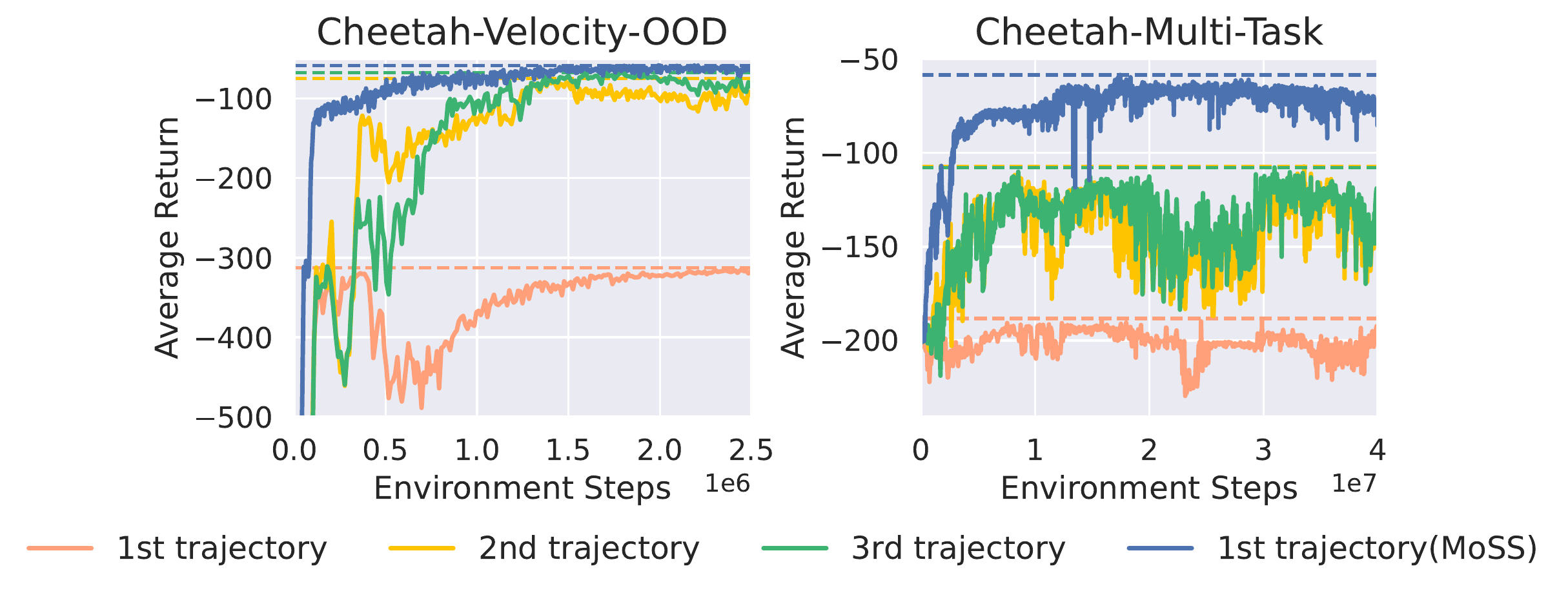}}
  \caption{Online adaptation ablation: MoSS updates the posterior task distribution at each time step so as to enable online task inference and adaptation. For comparison, we update the latent distribution at the trajectory level and report the first three episodes performance.}
	\label{fig:online-inference-ablation}
\end{figure*}
\label{fig:curve-ablation}
\begin{figure*}[t!]
  \centering
    \includegraphics[width=0.83\textwidth]{{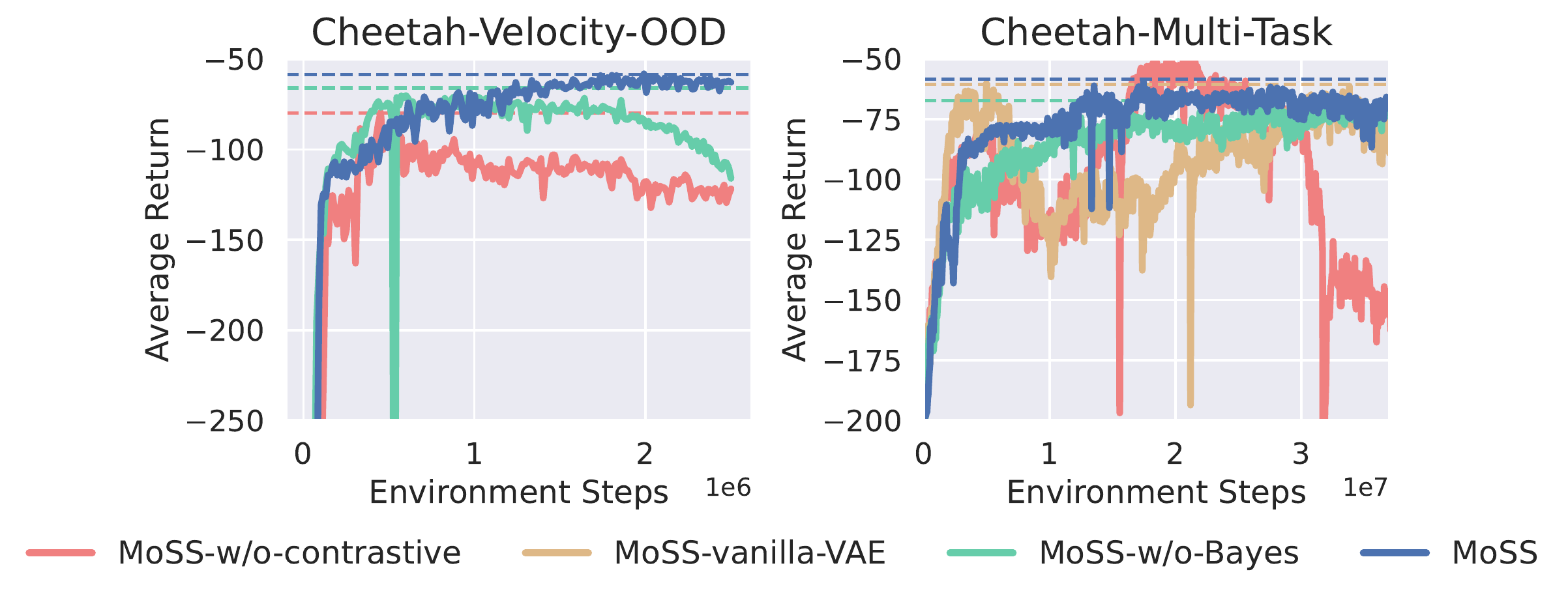}}
  \caption{MoSS ablation results: We compare the standard MoSS against ablation versions without Gaussian mixture latent space, without contrastive learning and without Bayes planning. The standard MoSS outperforms other methods and especially shows better stability.}
	\label{fig:curve-ablation}
\end{figure*}

\subsubsection{Online Task Inference Strategy}
\label{fig:latent-space-ablation}
\begin{figure}[hp]
  \centering
    \subfloat[MoSS]{\includegraphics[width=0.4\textwidth]{{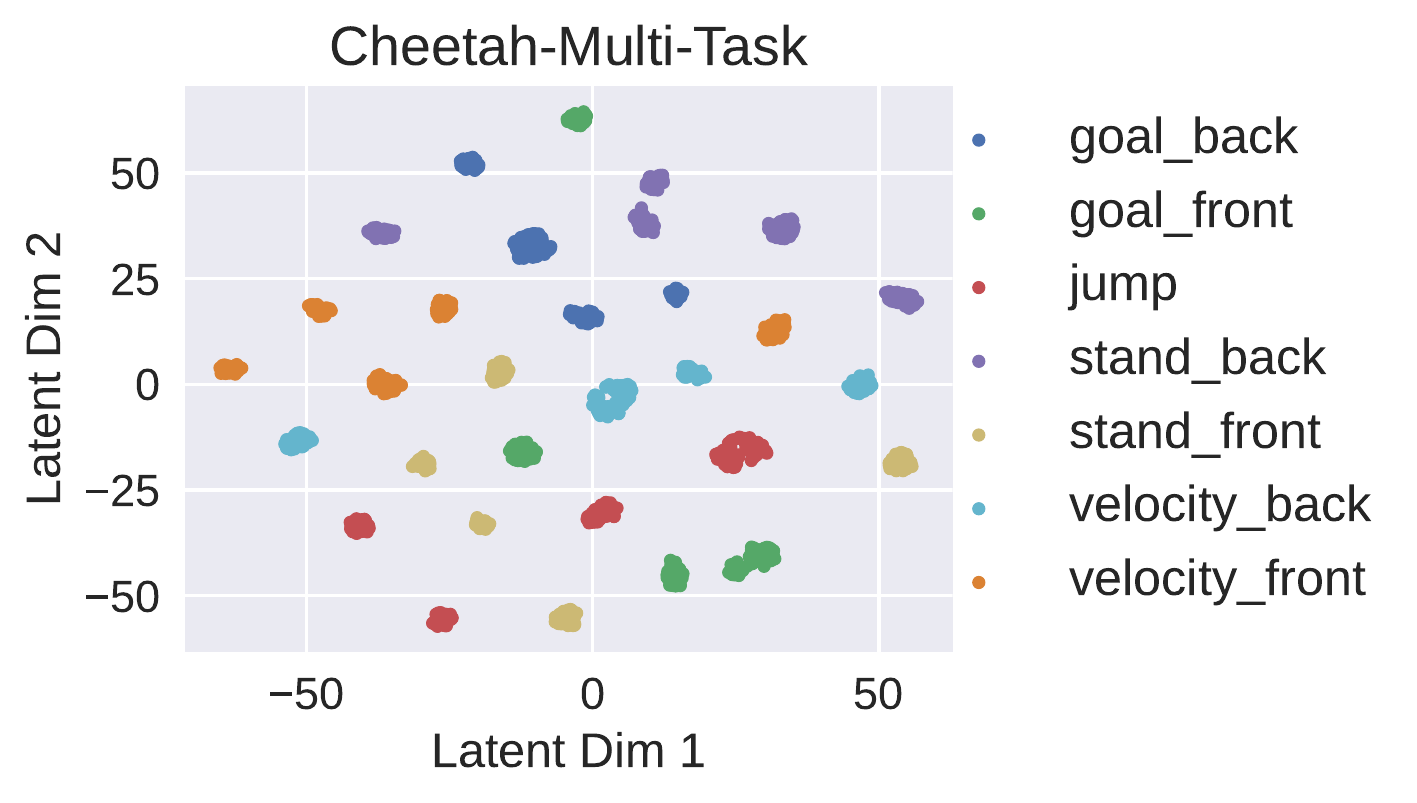}}}
    \\
    \subfloat[MoSS (single Gaussian)]{\includegraphics[width=0.4\textwidth]{{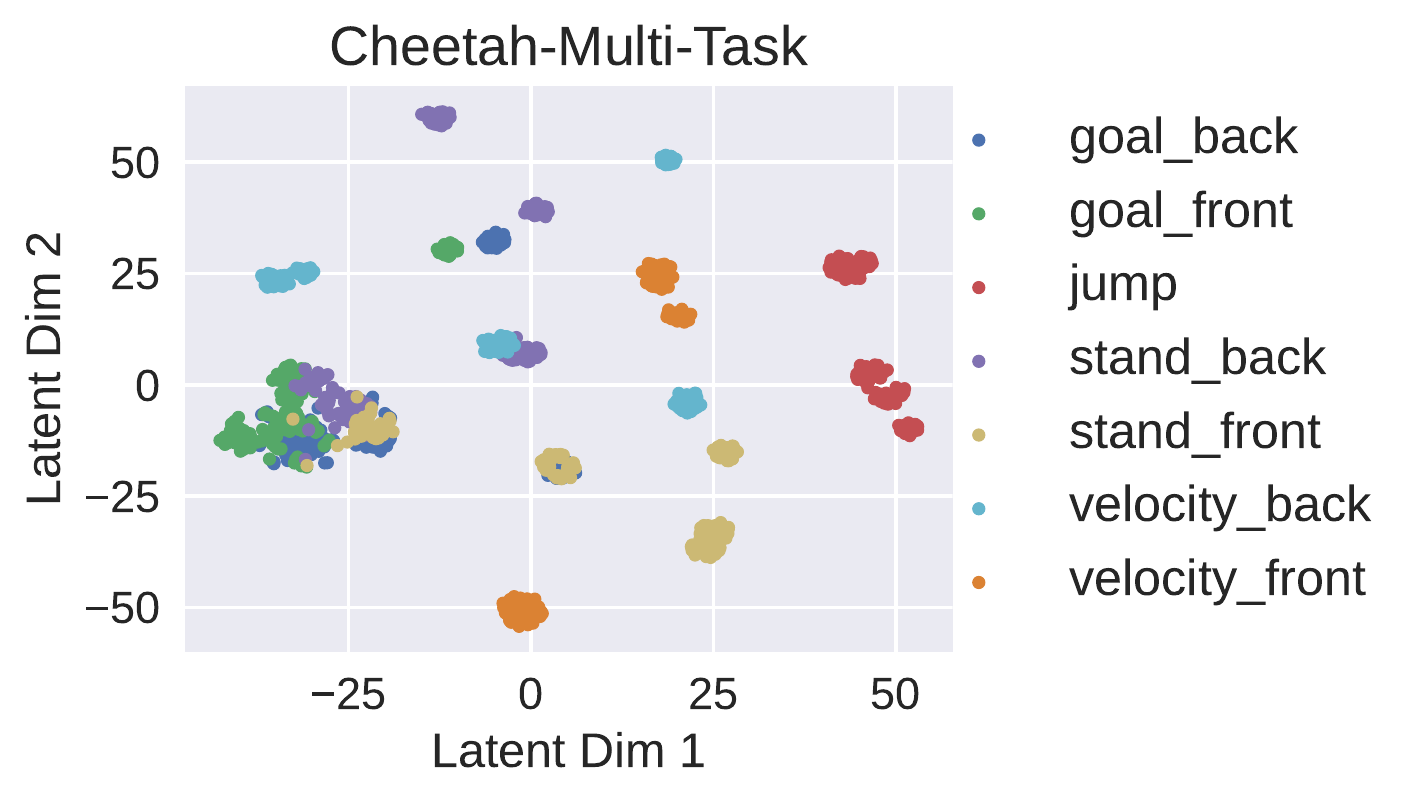}}}
    \\
    \subfloat[MoSS (no contrastive learning)]{\includegraphics[width=0.4\textwidth]{{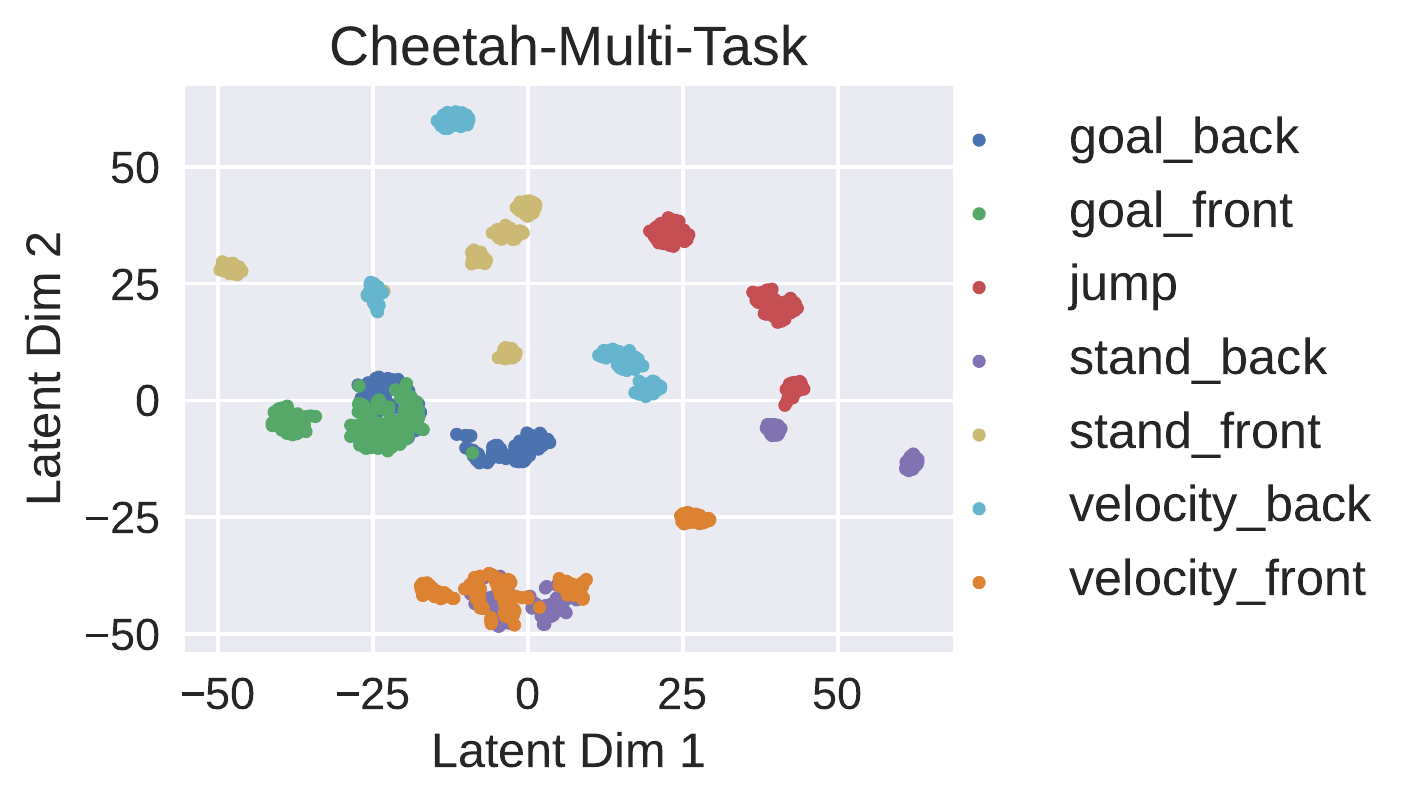}}}
  \caption{Latent embeddings ablation: We visualize the latent embedding space of MoSS and its ablation versions on \emph{Cheetah-Multi-Task}. (a) Latent task embeddings of the standard MoSS; (b) Latent task embeddings of MoSS with single Gaussian distribution for task representation; (c) Latent task embeddings of MoSS without contrastive learning objective. }
	\label{fig:latent-space-ablation}
\end{figure}

First, we investigate the efficacy of our online task inference mechanism. As an ablation version, we update the posterior task distributions at the trajectory level as in PEARL \cite{PEARL} rather than inferring tasks at each time step. The trajectory of the first episode is collected with prior distribution $p(\bm z) \sim \mathcal{N}(0, 1)$ and subsequent trajectories are collected with posterior task distributions $q(\bm z|\bm \tau)$ where the context $\bm \tau$ is aggregated over all past trajectories. In contrast, the standard MoSS updates its task representations based on latest context $q_{\bm \phi}(\bm z_{t}| \bm \tau_{:t}) = q_{\bm \phi}(\bm z_{t}| \bm s_{t}, \bm a_{t}, r_{t}, \bm h_{t-1})$ with $\bm h_{t-1}$ the GRU hidden state.

In Figure \ref{fig:online-inference-ablation}, we plot the average return of the first three episodes of the ablation model and compare it with the zero-shot result of standard MoSS. With online task inference and online adaptation, MoSS shows better performance in the first episode even with initial exploration steps, while MoSS with trajectory-wise adaptation requires additional exploration episodes to infer task information and act accordingly to solve the task. However, it still underperforms the standard MoSS, especially in the \emph{Cheetah-Multi-Task} environment. We suppose it is because the increased task difficulty imposes greater requirements on the agent's ability to reason about task information and it is insufficient to only capture rough task characteristics at the trajectory level. The comparison results demonstrate that MoSS is significantly more adaptation-efficient and shows better performance with the online inference strategy.

\subsubsection{Gaussian-Mixture Latent Space}
Second, we investigate the efficacy of our Gaussian mixture latent space for task representation. In the ablation study, we remove the Gaussian mixture strategy and use the standard VAE with single-component Gaussians to represent tasks. We compare it with standard MoSS on \emph{Cheetah-Multi-Task}. For a fair comparison, in the case of vanilla VAE, we use the latent dimension of 32, as in the standard MoSS we use the mixture of Gaussians with the dimension of 8 for each Gaussian component and a total of 4 Gaussian clusters. This ablation experiment is not conducted on \emph{Cheetah-Vel-OOD} as it only involves parametric variations and naturally uses the standard VAE as the inference module.

The comparison is given in Figure \ref{fig:curve-ablation} (right), although the average return of this ablation model increases rapidly at the beginning, it decreases afterwards and shows significant fluctuations. The reason could be that single-component Gaussian distributions are not sufficient to represent complex task relationships, and insufficient task inference makes policy learning difficult. This assumption is supported by the latent space visualization in Figure \ref{fig:latent-space-ablation} b, without explicitly classifying tasks as different base tasks, the task representations often overlap and thus it increases the difficulty of policy learning.

\subsubsection{Contrastive Learning}
Moreover, we ablate the contrastive learning objective of our algorithm. We run experiments on both \emph{Cheetah-Vel-OOD } and \emph{Cheetah-Multi-Task}. As shown in Figure \ref{fig:curve-ablation}, MoSS without contrastive learning is unstable and largely underperforms the original MoSS. We suppose it is also due to the insufficient task inference. Figure \ref{fig:latent-space-ablation} c shows the latent space structure generated by the model without contrastive learning. Only with reconstruction-based task inference training, the agent still cannot well differentiate context from distinct tasks, especially in the hard case as \emph{Cheetah-Multi-Task} where the task distribution includes both parametric and non-parametric variability. The inference network needs to distinguish not only the contexts that belong to tasks with different target values within the same base class, but also the contexts from different base tasks. In Figure \ref{fig:latent-space-ablation} c, task representations overlap especially between similar tasks like \emph{Back-Stand} and \emph{Jump}. However, the standard MoSS still handles the task relationships well and produces a well-structured latent space that benefits further policy learning.

\subsubsection{Bayes-Adaptive Planning}
Finally, we compare two strategies for task representation in the policy module. In MoSS, we adopt the idea of Bayes-adaptive planning \cite{BAMDP, BayesianRL} and condition the policy $\pi_{\bm \theta}(\bm a|\bm s, q_{\bm \phi}(\bm z))$ on both the agent state $\bm s_{t}$ and the latent task distribution $q_{\bm \phi}(\bm z)$ to explicitly accommodate uncertainty in the action selection process. Another strategy as used in \cite{PEARL} is to perform posterior sampling $\bm z \sim q_{\bm \phi}(\bm z)$ and condition the policy $\pi_{\bm \theta}(\bm a|\bm s, \bm z)$ on sampled $\bm z$ rather than the whole distribution. In Bayes-adaptive planning, the agent’s uncertainty about the environment is directly fed into the policy network so that a Bayes-optimal agent systematically seeks out actions needed to quickly reduce uncertainty and also helps maximise expected return. While posterior sampling reduces the meta-RL planning problem into a regular MDP by periodically sampling a hypothesis MDP from its posterior. 
MoSS without Bayes adaptive planning still has good stability and generally achieves comparable but slightly worse performance than the standard MoSS. As we update the posterior task belief $q_{\bm \phi}(\bm z_{t})$ at every time step, the Bayes-adaptive planning might not have a big impact on the overall performance: although the posterior sampling at some time step induces some inaccuracy, at the next time step the model will update the task belief and perform sampling again, thus naturally there is some tolerance for inaccuracy.

\subsubsection{MuJoCo Experiments}
\subsubsection{Complete experiment results in parametric MuJoCo environments}
In Section \ref{fig:mujoco-parametric-result}, we present the experiment results on four environments: \textit{Cheetah-Vel}, \textit{Cheetah-Dir}, \textit{Ant-Dir} and \textit{Walker-Rand-Params}. Prior works \cite{PEARL, VariBAD} also use another two environments \textit{Ant-Goal} and \textit{Humanoid-Dir} to evaluate the meta-RL algorithm. Here we also include these environments and present the results in six parametric MuJoCo environments in Figure \ref{fig:complete-parametric}.

\label{fig:complete-parametric}
\begin{figure*}[t]
  \centering
    \includegraphics[width=0.9\textwidth]{{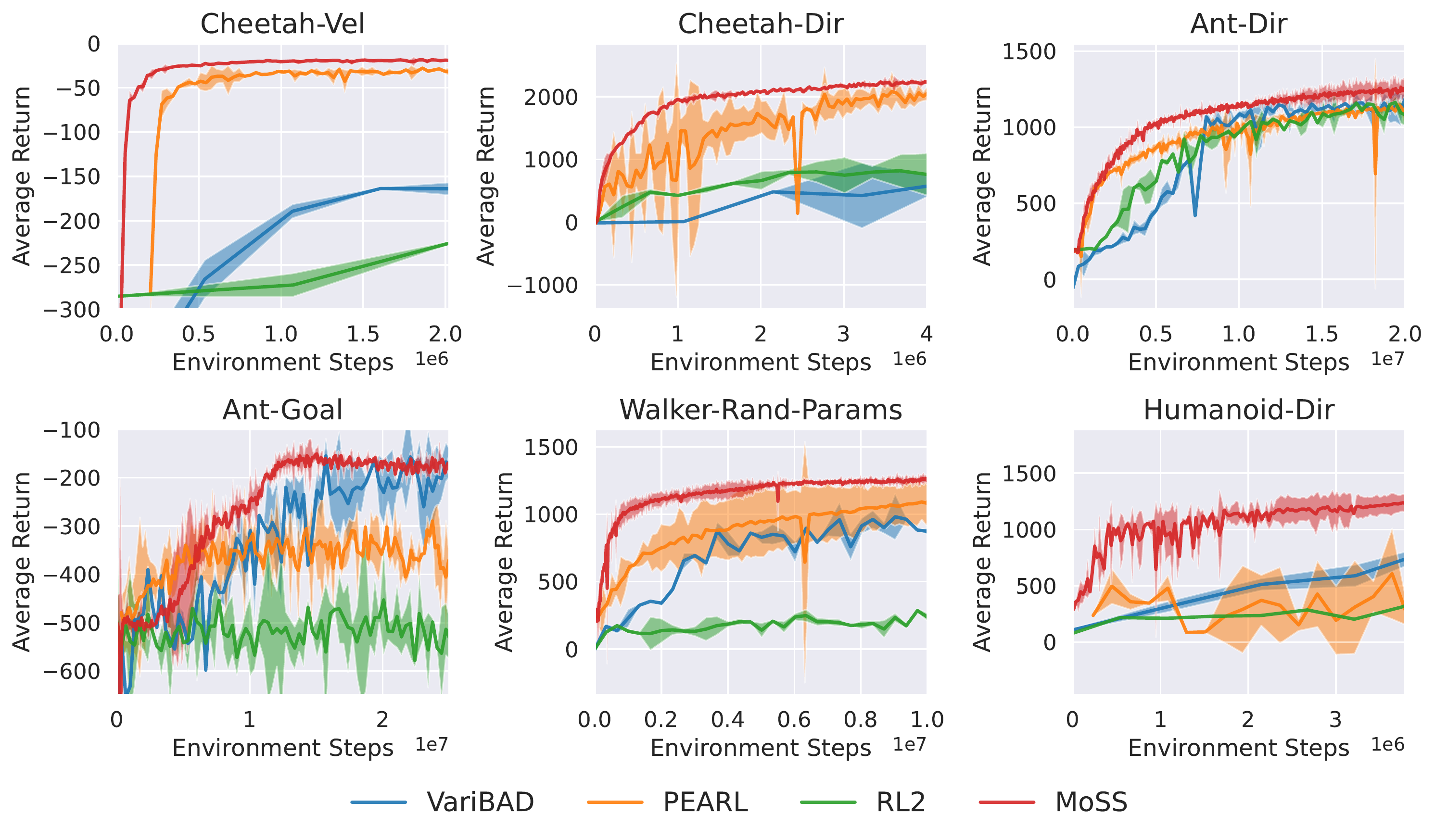}}
  \caption{Complete meta-test performance in parametric MuJoCo environments}
    \label{fig:complete-parametric}
\end{figure*}

\subsubsection{Agent response in non-stationary environments}
\label{sec:agent-response-appendix}
In Figure \ref{fig:non-stationary-result}, we plot the agent responce to task changes in the \textit{Cheetah-Direction-Non-Stat} environment. In Figure \ref{fig:complete-agent-response}, we give more results in all three non-stationary environments.

\label{fig:complete-agent-response}
\begin{figure*}[t!]
  \centering
    \includegraphics[width=0.3\textwidth]{{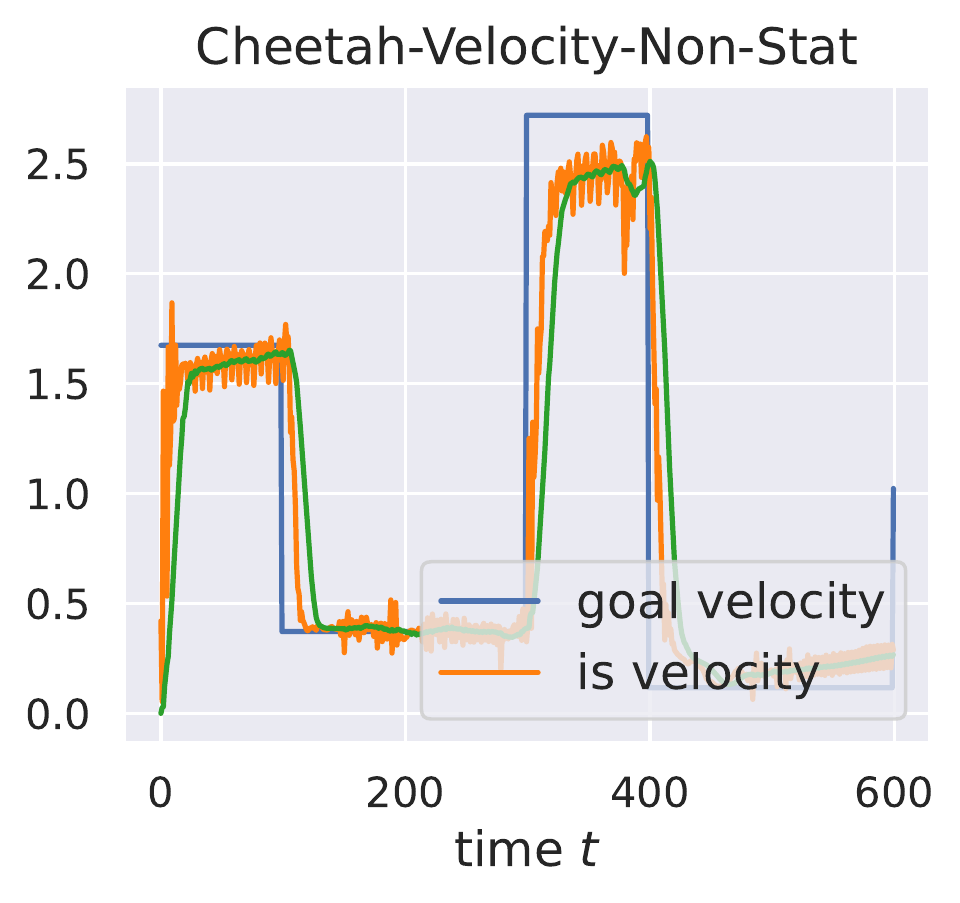}}
    \includegraphics[width=0.3\textwidth]{{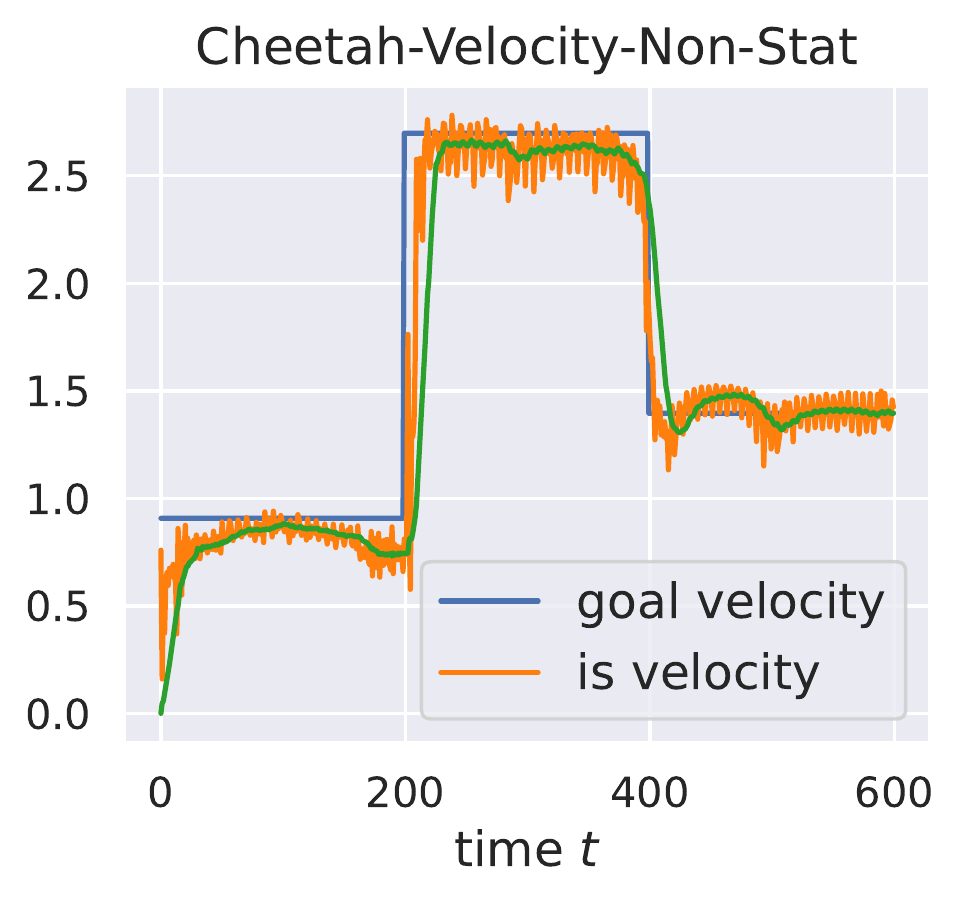}}
    \includegraphics[width=0.3\textwidth]{{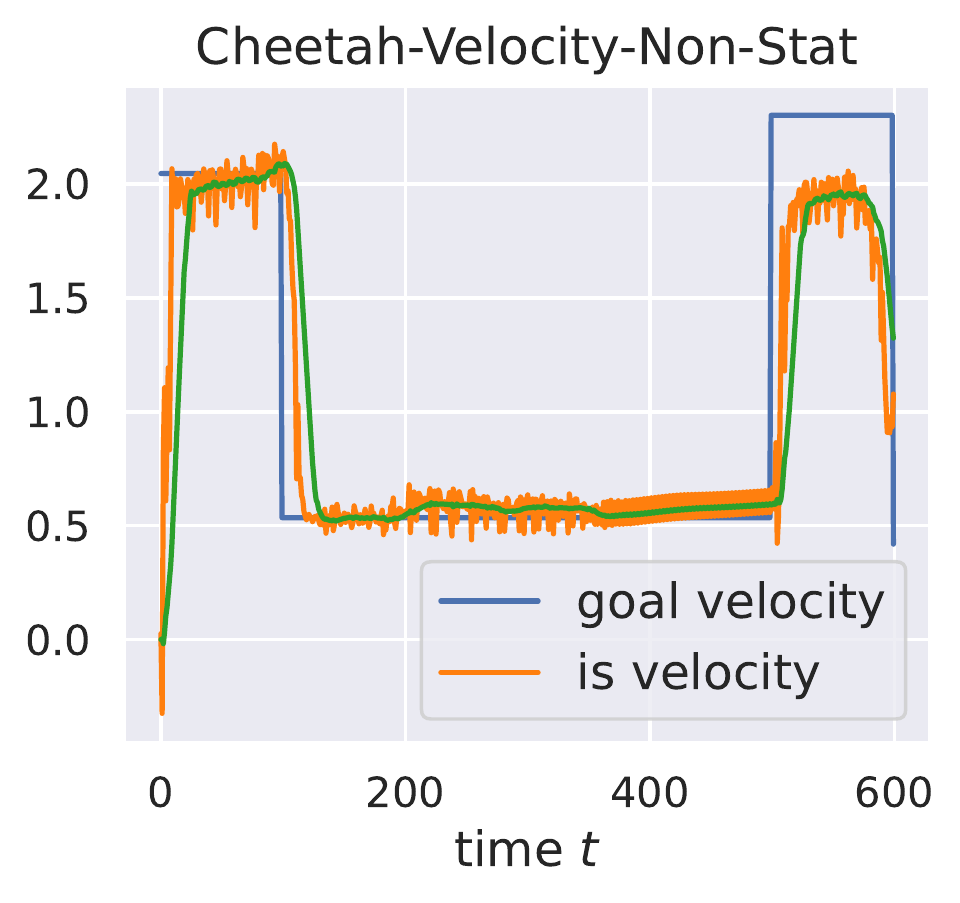}}
    \includegraphics[width=0.3\textwidth]{{results/response-cheetah-dir.pdf}}
    \includegraphics[width=0.3\textwidth]{{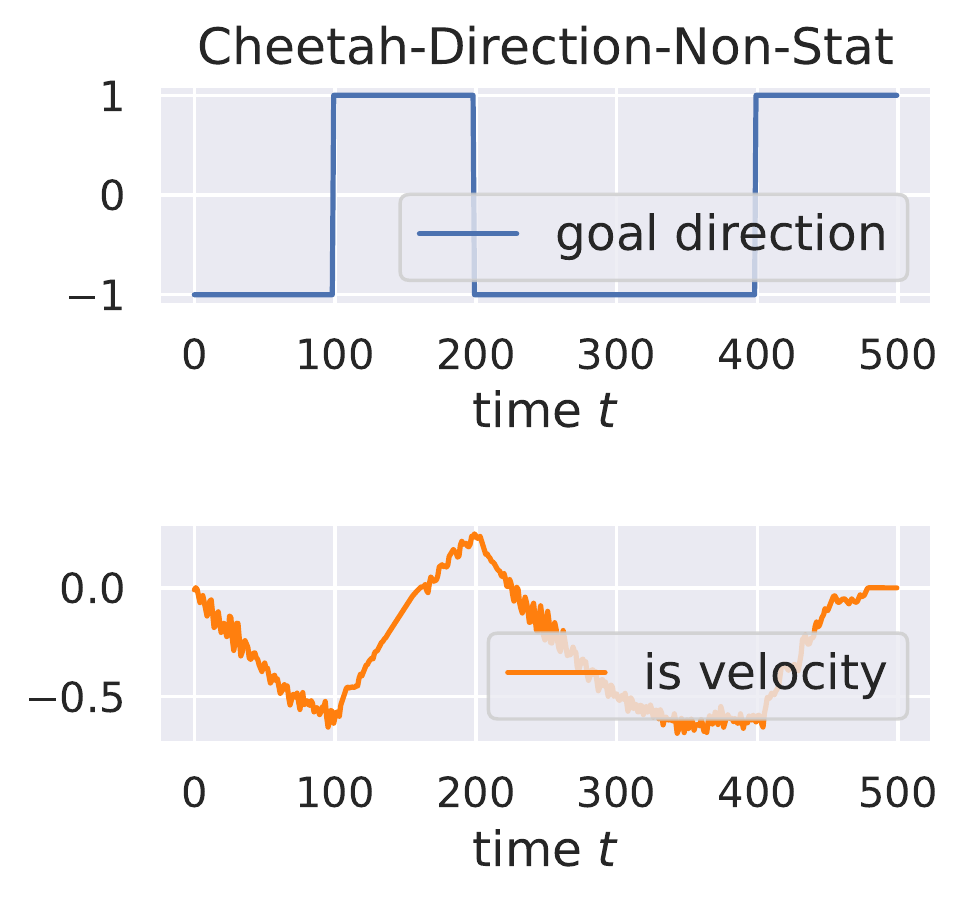}}
    \includegraphics[width=0.3\textwidth]{{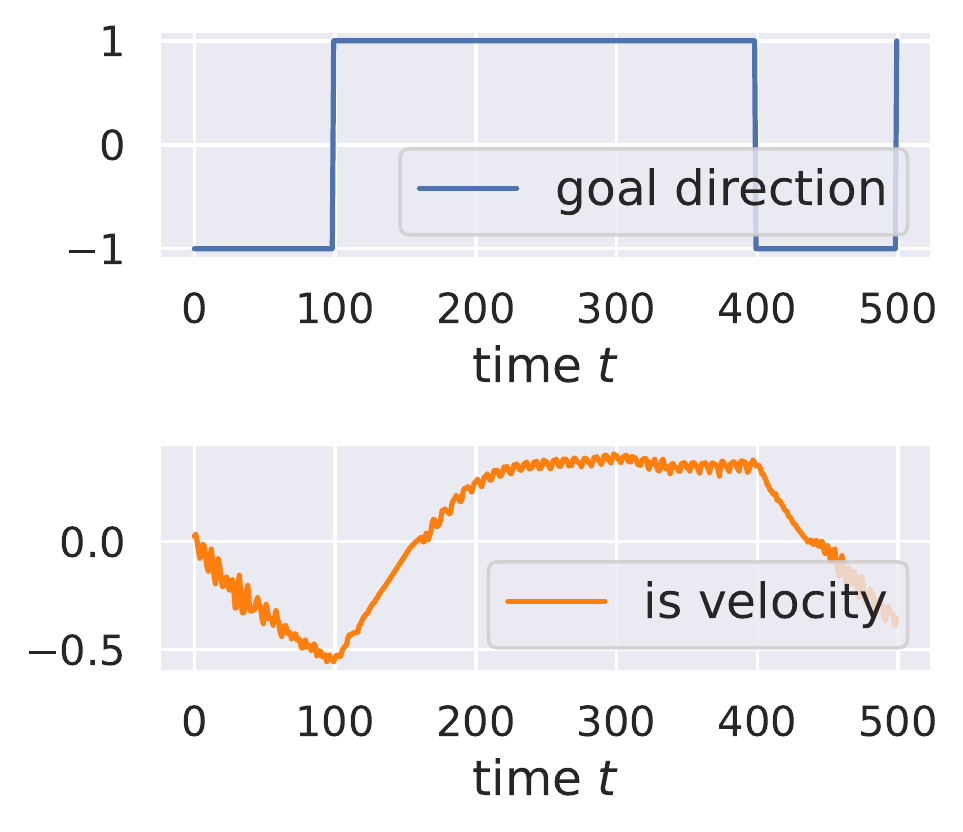}}
    \includegraphics[width=0.3\textwidth]{{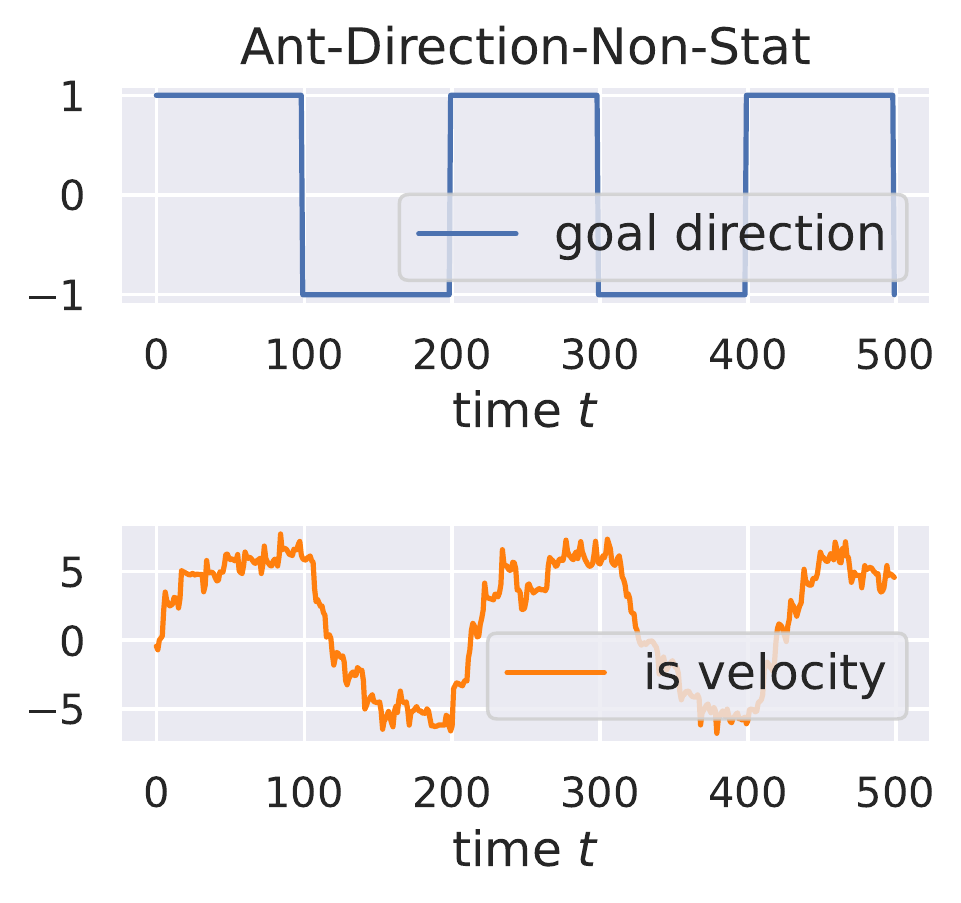}}
    \includegraphics[width=0.3\textwidth]{{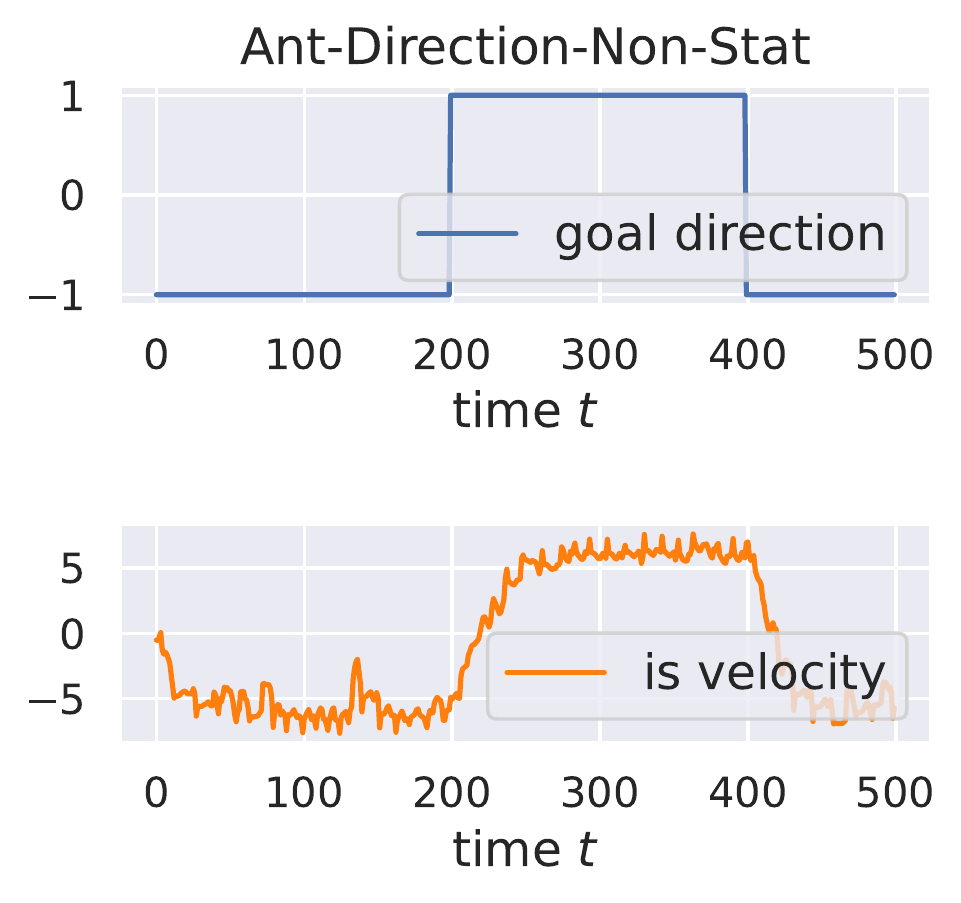}}
    \includegraphics[width=0.3\textwidth]{{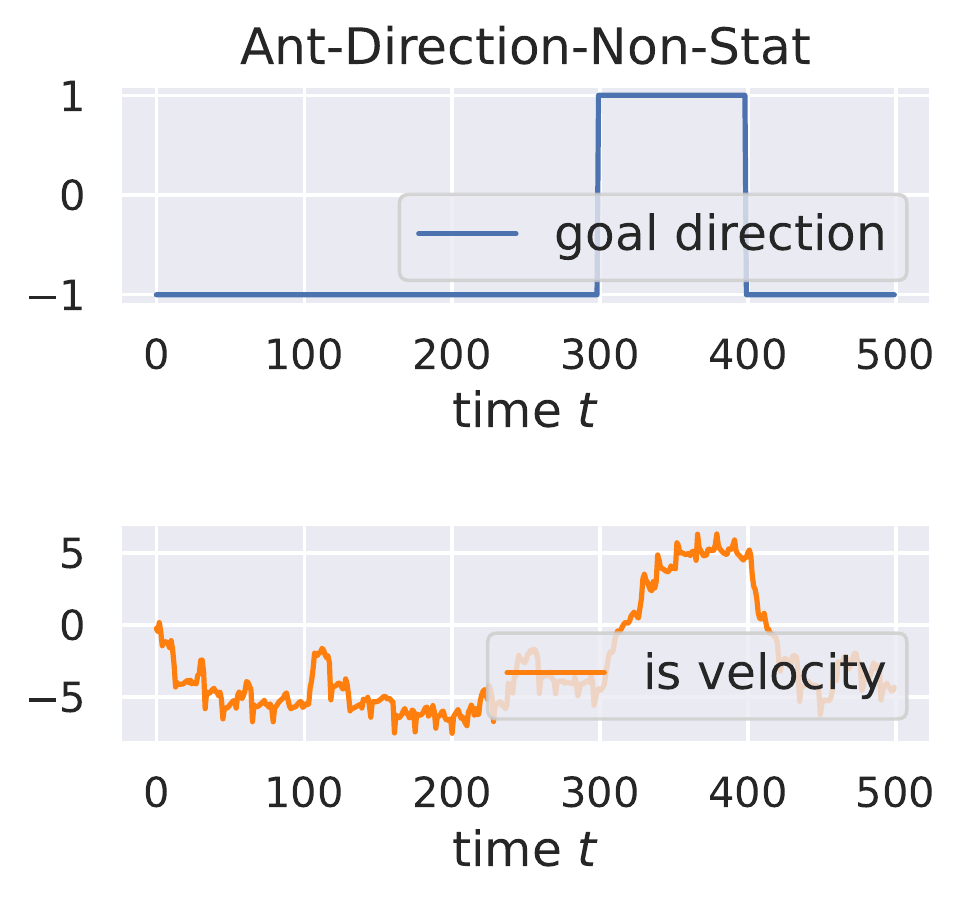}}
  \caption{Agent responses to task changes in non-stationary MuJoCo environments: We change the goal value (velocity or direction) randomly at every 100 time step at the probability of 0.5 and visualize the agent response to task changes. As shown, the MoSS agent shows good tracking behavior, it can detect the task change and quickly adapt to the new task.}
	\label{fig:complete-agent-response}
\end{figure*}

\subsubsection{Meta-World Experiments}
\label{sec:meta-world-appendix}

\subsubsection{Meta-World ML1 V1 Results}
\label{sec:meta-world-v1v2}
As mentioned in Section \ref{sec:parametric-result}, we mainly evaluate MoSS on the Meta-World ML1 V2 benchmark. In this new version of Meta-World, the environment settings and the reward function design have been
modified compared to its original version (Meta-World-V1). However, VariBAD \cite{VariBAD}
only uses the original version, Meta-World-V1, the reward function design of which is sub-optimal as
discussed in the updated paper of Meta-World \cite{MetaWorld} and its open-source repository\footnote{\url{https://github.com/rlworkgroup/metaworld}}. Nevertheless, for a fair comparison, we also run experiments on Meta-World-V1, results of success rates are summarized in Table \ref{tab:meta-world-v1v2-result}. For training curves, see Figure \ref{fig:meta-world-training-curves}.

\footnotetext{Baseline results are taken from \cite{MetaWorld} as the performance is highly sensitive to hyperparameters and hard to reproduce.}

MoSS still shows comparable performance on Meta-World V1. However, it fails to achieve a high success rate on the \emph{Reach} task, which is somehow counter-intuitive, as \emph{Reach} is the easiest among the three tasks and can be seen as the early-stage sub-task of the \emph{Push} or \emph{Pick-and-Place} task. We hypothesize that it might be the sub-optimal reward function design that leads to the problem. Also, the Meta-World benchmark is very sensitive to the hyperparameter choice, inappropriate hyperparameters could be another reason for the low success rate on the \emph{Reach} task.

\begin{table*}[tp]
  \centering
  \caption {Meta-World ML 1 V1 and V2 results}\footnotemark
    \begin{tabular}{c|c|c|c|c|c|c|c}
    \toprule
    \multirow{2}[4]{*}{Method} & \multirow{2}[4]{*}{$k$-th episode} & \multicolumn{3}{c|}{Meta-World V1 (\%)} & \multicolumn{3}{c}{Meta-World V2 (\%)} \\
\cmidrule{3-8}          &       & Reach & Push  & Pick-Place & Reach & Push  & Pick-Place \\
    \midrule
    MAML  & 10    & 48    & 74    & 12    & \textbf{100} & 94    & 80 \\
    RL$^{2}$   & 10    & 45    & 87    & 24    & \textbf{100} & 96 & 98 \\
    PEARL & 10    & 38    & 71    & 28    & 68    & 44    & 28 \\
    VariBAD & 1 or 2   & \textbf{100} & \textbf{100} & 29    & -     & -     & - \\
    MoSS (ours) & \textbf{1}     & 52    & \textbf{100} & \textbf{60} & 86    & \textbf{100}    & \textbf{100} \\
    \bottomrule
    \end{tabular}
  \label{tab:meta-world-v1v2-result}
\end{table*}

\subsubsection{Meta-World training curves}
\label{sec:meta-world-curve}
Meta-World uses success rate as its evaluation metric, as values of reward do not directly indicate how successful a policy is \cite{MetaWorld}. We show the training curves of MoSS on Meta-World ML1 tasks in Figure \ref{fig:meta-world-training-curves}. We only train MoSS on Meta-World and other results given in Table \ref{tab:meta-world-result} and Table \ref{tab:meta-world-v1v2-result} are taken from \cite{MetaWorld, VariBAD}, as we cannot reproduce their reported results due to the sensitivity of Meta-World tasks to hyperparameters.

\label{fig:meta-world-training-curves}
\begin{figure*}[tp]
  \centering
  \subfloat[][Meta-World V1 training curves]{
   \includegraphics[width=0.7\textwidth]{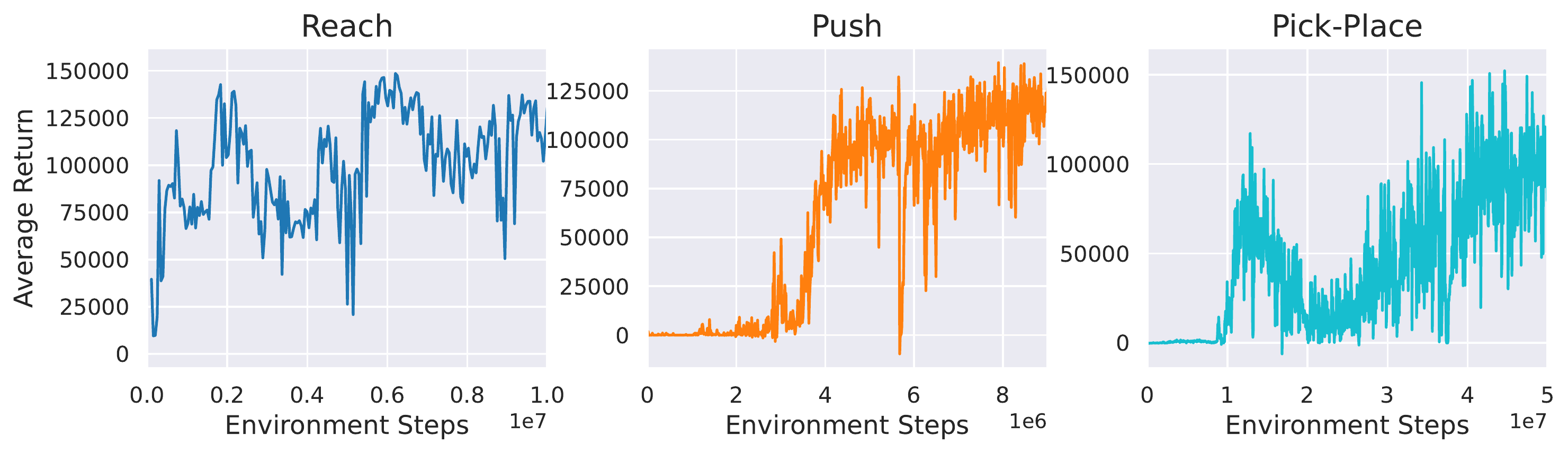}
   }
   \\
    \subfloat[][Meta-World V2 training curves]{
   \includegraphics[width=0.7\textwidth]{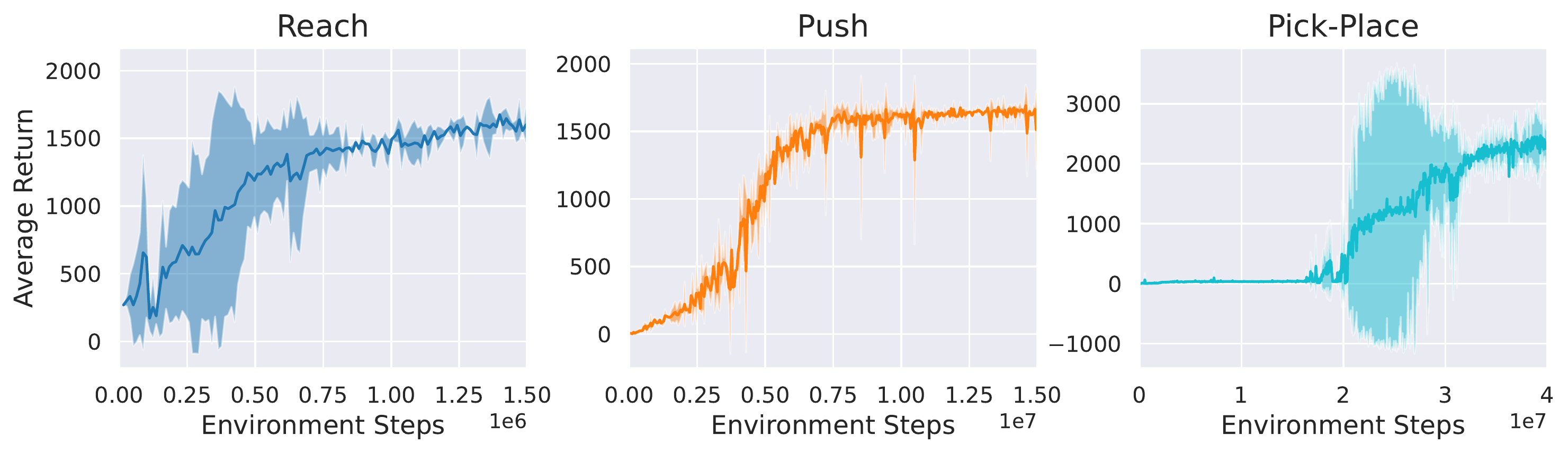}
   }
  \caption{Training curves of MoSS on Meta-World V1 and V2 benchmarks: Average return (y-axis) against collected environment steps during meta-training (x-axis).}
	\label{fig:meta-world-training-curves}
\end{figure*}

As shown in Figure \ref{fig:meta-world-training-curves}, average returns are more stabler on Meta-World V2 because of better reward function design. This indicates the problem of inappropriate reward function design of Meta-World V1 on the one hand, and the dependence of the algorithm's performance on reward design on the other hand. It is worth studying how to make the meta-RL algorithm less dependent on the reward design and even makes it applicable in sparse-reward settings in future work.

\subsubsection{Full timescale Results}
\label{sec:full-timescale-results}

In Section \ref{sec:experiments}, we truncate the x-axis to better illustrate the performance of MoSS.
Here, we plot MoSS and PEARL against the on-policy methods that run for
the full time steps (1e8). MoSS shows superior sample efficiency in all experiments with diverse task distributions. We show the results of parametric and out-of-distribution tasks in Figure \ref{fig:full-timescale-parametric-n-ood}, results of non-stationary tasks in Figure \ref{fig:full-timescale-non-stationary}, and results of non-parametric tasks in Figure \ref{fig:full-timescale-non-parametric}, respectively.
\label{fig:full-timescale-parametric-n-ood}
\begin{figure*}[tp]
  \centering
    \includegraphics[width=1\textwidth]{{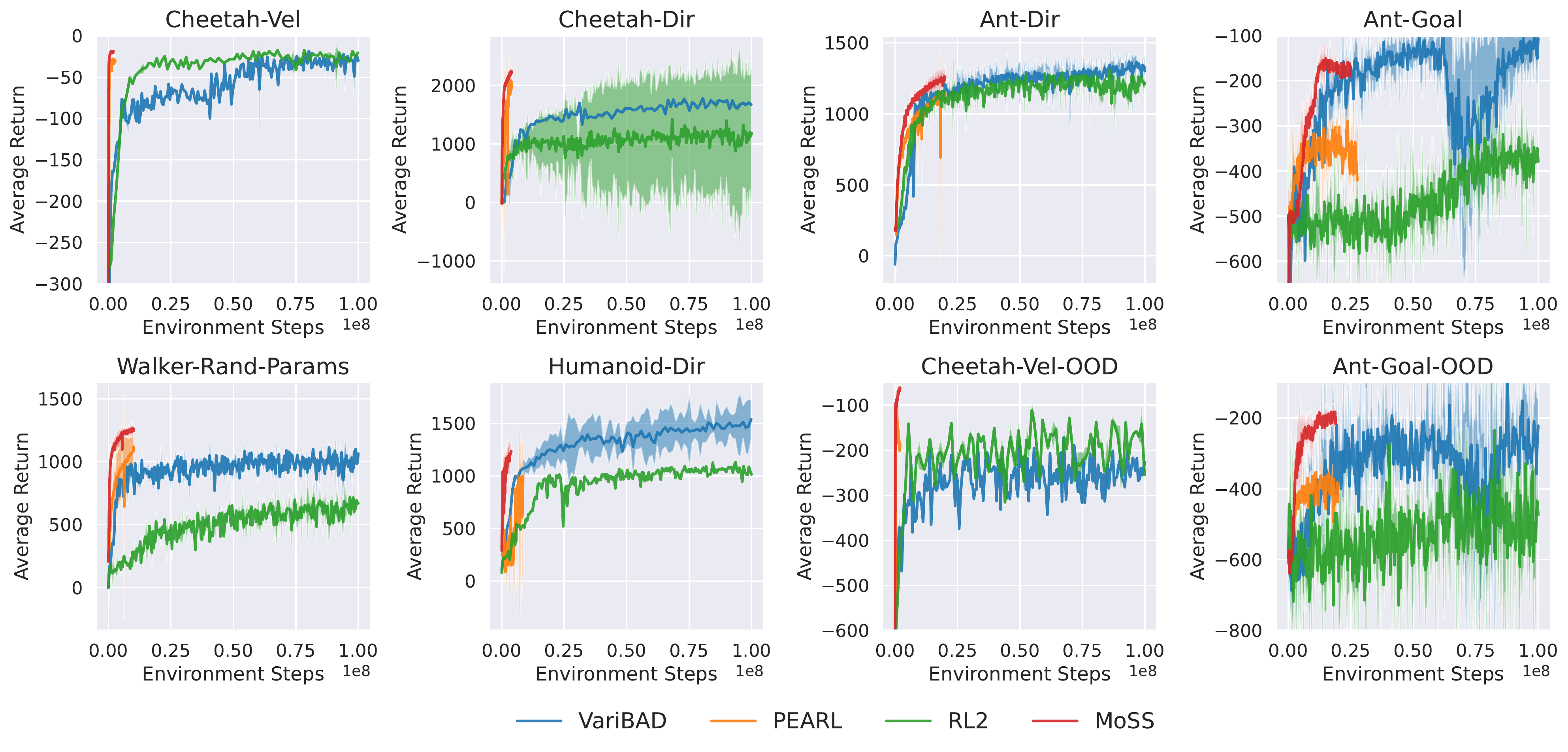}}
  \caption{Full timescale results in parametric MuJoCo environments with in-distribution and out-of-distribution test tasks}
	\label{fig:full-timescale-parametric-n-ood}
\end{figure*}

\label{fig:full-timescale-non-stationary}
\begin{figure*}[t]
  \centering
    \includegraphics[width=0.8\textwidth]{{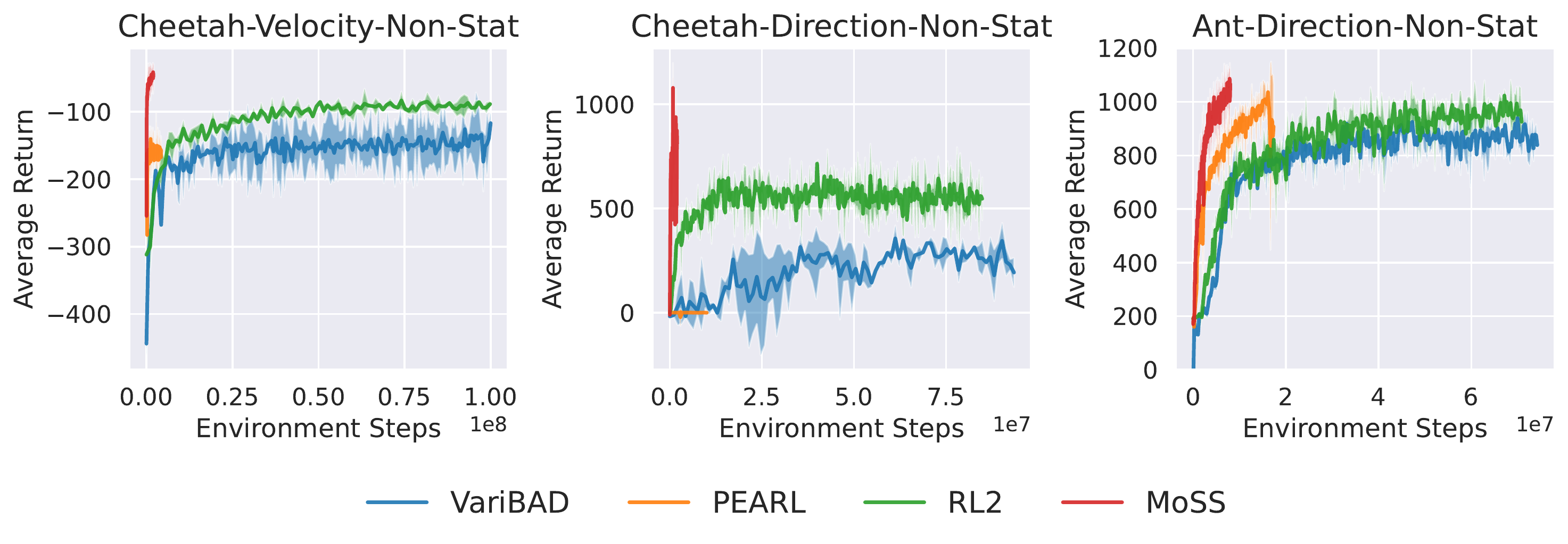}}
  \caption{Full timescale results in non-stationary MuJoCo environments}
	\label{fig:full-timescale-non-stationary}
\end{figure*}

\label{fig:full-timescale-non-parametric}
\begin{figure*}[t]
  \centering
    \includegraphics[width=0.6\textwidth]{{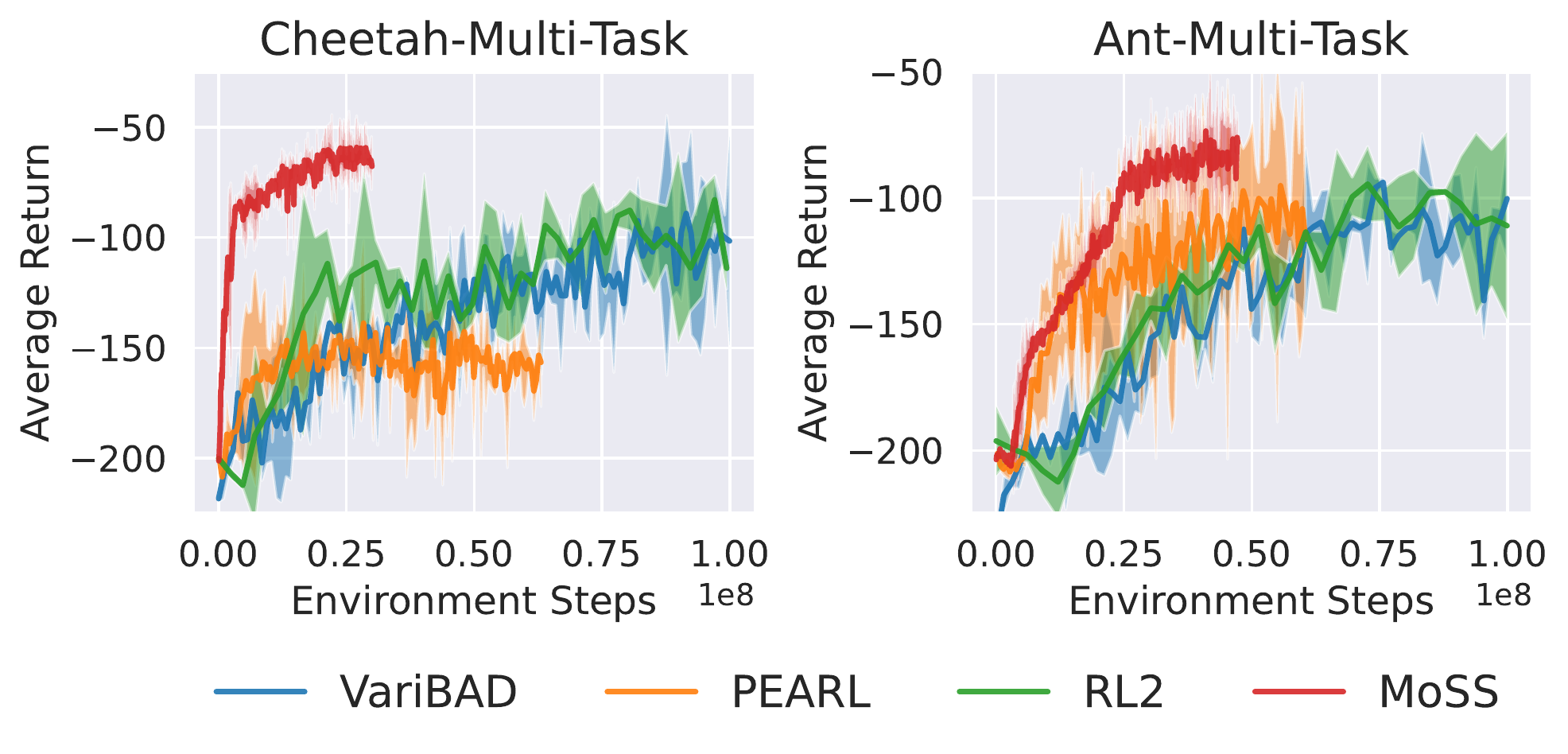}}
  \caption{Full timescale results in non-parametric MuJoCo environments}
	\label{fig:full-timescale-non-parametric}
\end{figure*}

\begin{table*}[t!]
  \centering
  \caption{\emph{Cheetah-Multi-Task} benchmark description}
  \renewcommand\arraystretch{1.5}
    \begin{tabular}{c|c|c|c}
    \toprule
    Base Task & Behavior & Range & Reward \\
    \bottomrule
    Velocity & Run at goal velocity\  (forward/backward) & $|v_{x}^{*}| \in [1,\  5]$ & $r = -|v_{x}^{*} -v_{x}|$ \\
    \hline
    Goal  & Reach goal position\  (forward/backward) & $|p_{x}^{*}| \in [5,\  25]$ & $r = -|p_{x}^{*} -p_{x}|$ \\
    \hline
    Stand & Stand at goal angle\  (forward/backward) & $|p_{y}^{*}| \in [\frac{\pi}{6},\  \frac{\pi}{2}]$ & $r = -|p_{y}^{*} -p_{y}|$ \\
    \hline
    Jump  & Jump upwards & $|v_{z}^{*}| \in [1.5,\  3.0]$ & $r = -|v_{z}^{*} -|v_{z}||$ \\
    \bottomrule
    \end{tabular}%
  \label{tab:cheetah-multi-task-description}%
\end{table*}%

\begin{table*}[t!]
  \centering
  \caption{\emph{Ant-Multi-Task} benchmark description}
  \renewcommand\arraystretch{1.5}
    \begin{tabular}{c|c|c|c}
    \toprule
    Base Task & Behavior & Range & Reward \\
    \bottomrule
    Velocity & Run at goal velocity (up, down, left, right) & $|v_{x}^{*}| \in [1,\  3]$ & $r = -|v_{x}^{*} -v_{x}|$ \\
    \hline
    Goal  & Reach goal position (up, down, left, right) & $|p_{x}^{*}| \in [5,\  15]$ & $r = -|p_{x}^{*} -p_{x}|$ \\
    \hline
    Jump  & Jump upwards & $|v_{z}^{*}| \in [0.5,\  2.0]$ & $r = -|v_{z}^{*} -|v_{z}||$ \\
    \bottomrule
    \end{tabular}%
  \label{tab:ant-multi-task-description}%
\end{table*}%

\begin{table*}[t!]
  \centering
  \caption{MoSS comparison with baseline methods}
    \begin{tabular}{c|c|c|c|c|c|c}
    \toprule
    \multicolumn{1}{p{4em}|}{Method} 
    &
    \multicolumn{1}{p{4.5em}|}{\makecell[c]{Gaussian\\ mixture}} 
    &
    \multicolumn{1}{p{4.7em}|}{\makecell[c]{Contrastive \\ loss}} 
    &
    \multicolumn{1}{p{4.5em}|}{\makecell[c]{Bayes \\Adaptive}} 
    &
    \multicolumn{1}{p{4.5em}|}{\makecell[c]{Off-policy\\ RL}} 
    &
    \multicolumn{1}{p{4.5em}|}{\makecell[c]{Off-policy \\inference}} 
    &
    \multicolumn{1}{p{4.5em}}{\makecell[c]{Online \\Inference}} \\
    \midrule
    \textbf{MoSS} & \textbf{Yes} & \textbf{Yes} & \textbf{Yes} & \textbf{Yes} & \textbf{Yes} & \textbf{Yes} \\
    RL$^{2}$   & No    & No    & No    & No    & No    & \textbf{Yes} \\
    PEARL & No    & No    & No    & \textbf{Yes} & No    & No \\
    VariBAD & No    & No    & \textbf{Yes} & No    & \textbf{Yes} & \textbf{Yes} \\
    \bottomrule
    \end{tabular}%
  \label{tab:methods-comparison}%
\end{table*}%

\subsection{Implementation Details}
\label{sec:implementation-details}
\subsubsection{MoSS task inference details}
\label{sec:moss-model-details}

In this section, we provide detailed description and illustration (see Figure \ref{fig:moss-task-inference}) of the task inference module of MoSS.
\label{fig:moss-task-inference}
\begin{figure}[t]
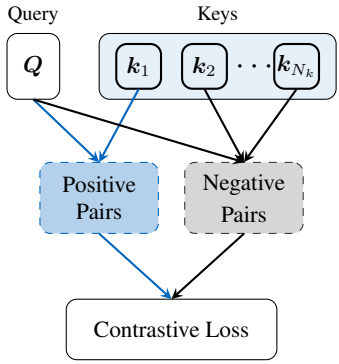

    \centering
    \drawcontrastive
        \caption{We use the latent task embeddings as queries and keys, where embeddings from the same trajectory (but of different time steps) are positive pairs and embeddings from trajectories of different tasks are negative pairs. Contrastive learning aims to identify positive sample pairs amongst a set of negative pairs.
    }
    \label{fig:moss-task-inference}
\end{figure}

\subsubsection{Gaussian mixture latent space}
As discussed in Section \ref{sec:gaussian-mixture}, we extend the VAE network with the Gaussian mixtrue latent space to accommodate parametric and non-parametric task variability. Specifically, we use a categorical variable $\bm y \sim p(\bm y) = \text{Cat}(\pi) $ to represent the base task probability distribution, and use Gaussian component $\mathcal{N}(\mu_{\bm z}(\bm y), \sigma_{\bm z}(\bm y))$ to represent parametric task variability within each base task. Given the observation $\bm x$ and latent variables $\bm y, \bm z$, we use the following generative model as in \cite{VaDE, CURL-Rao}:

\begin{itemize}
\item 1. Choose a cluster $\bm y \sim p(\bm y) = Cat(\pi)$
\item 2. Choose a latent variable $\bm z \sim p(\bm z | \bm y) = \mathcal{N}(\bm \mu_{z}(\bm y), \bm \sigma_{z}^{2}(\bm y))$ \par
\item 3. Decode to reconstruct the observation $\bm x$: $\bm x \sim p(\bm x|\bm z)$
\end{itemize}

The categorical variable $\bm y$ indicates which base task the input task trajectory $\bm x$ belongs to, denoted as a one-hot vector. Then given a certain class $\bm y(k)$, the latent variable $\bm z$ is drawn based on the task-specific Gaussian distribution $\mathcal{N}(\mu_{\bm z}(\bm y), \sigma_{\bm z}(\bm y))$. Therefore, latent variables $\bm z$ are formulated as a mixture of Gaussians, which encodes both the inter- and intra-cluster variations.

\label{fig:moss-decoder}
\begin{figure}[tp]
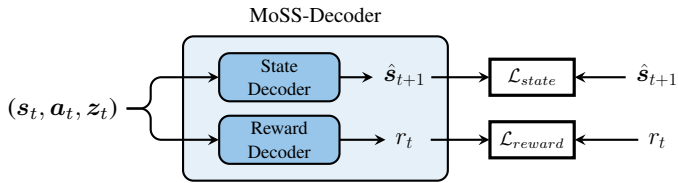

  \centering
  \drawdecoder
  \caption{
    We parameterize the decoder as two independent networks: the state decoder and the reward deocder. We use the reconstruction loss between the prediction $(\bm \hat{s}_{t+1}, \hat{r}_{t})$ and the true target $(\bm s_{t+1}, r_{t})$ to optimize the encoder-decoder network.}
	\label{fig:moss-decoder}
\end{figure}

The decoder maps from the latent variable $\bm z$ to reconstructed data $\bm \hat{x}$. We use the reconstruction loss between the input $\bm x$ and the prediction $\bm \hat{x}$ to optimize the task inference module. As we use the input data itself as the supervisory label, the task inference optimization belongs to a self-supervised approach and does not require any privileged task information. As shown in Figure \ref{fig:moss-decoder}, we parameterize the decoder as two independent regression networks: the state decoder $p_{\bm \phi}(\bm s_{t+1}|\bm s_{t}, \bm a_{t}, \bm z_{t})$ and the reward decoder $p_{\bm \phi}(\bm r_{t}|\bm s_{t}, \bm a_{t}, \bm z_{t})$. We use the mean squared error (MSE) to calculate the prediction loss. Therefore, the reconstruction term $p(\bm x|\bm z_{k})$ in Equation \ref{con:ELBO} is factorized as:

\begin{equation}
\begin{aligned}
\log p(\bm x_{t}|\bm z_{t}) &= \log p_{\bm \phi}(\bm s_{t+1}, r_{t}|\bm s_{t}, \bm a_{t}, \bm z_{t}) \\
&= \log p_{\bm \phi}(\bm s_{t+1}|\bm s_{t}, \bm a_{t}, \bm z_{t}) + \log p_{\bm \phi}(r_{t}|\bm s_{t}, \bm a_{t}, \bm z_{t}) \\
& \approx \|\hat{\bm s}_{t+1}-\bm s_{t+1}\|^{2}+(\hat{\bm r}_{t}-r_{t})^{2}
\end{aligned}
\end{equation}

\subsubsection{Contrastive learning}

We expect the encoder to distinguish different tasks by organizing their latent embeddings in a structured way, i.e., the latent embeddings of the same task should stay close to each other, while embeddings from different tasks should be differentiable. Therefore, we also introduce the contrastive loss to jointly train the task inference module, so that the model can better capture task characteristics and produces more reliable task representations. The calculation of contrastive loss is illustrated in Figure \ref{fig:contrastive-learning}. As mentioned in Section \ref{sec:contrastive-learning}, we use the latent task embeddings as queries and keys. For each query embedding $\bm Q$, we sample one latent embedding from the same trajectory (but of different time steps) as the positive key, and $N_k - 1$ embeddings from trajectories of different tasks as negative keys. Given positive and negative pairs, we use the InfoNCE score \cite{InfoNCE} (Equation \ref{con:infoNCE}) to calculate the contrastive loss and optimize the encoder-decoder network jointly with the VAE reconstruction loss. Note that We don’t use contrastive learning for non-stationary tasks as latent embeddings from the same episode but different time steps could also be positive pairs. Even without contrastive loss MoSS still perform pretty well in such cases, detailed results are shown in \ref{sec:non-stationary results} and \ref{sec:agent-response-appendix}.

\textcolor{blue}{\subsubsection{Comparison with baseline methods}}
Here we compare several features of MoSS with RL2, PEARL and VariBAD to clarify the difference between MoSS and other baseline methods. The comparison is given in Table \ref{tab:methods-comparison}

\subsubsection{Environmental details}
\label{sec:environmental-details}

Here we describe the meta-learning domain of each environment used in this work.\par

\subsubsection{MuJoCo parametric environments} \par
\label{sec:mujoco-parametric-details}
\begin{itemize}
\item Cheetah-Vel: achieve a target velocity $ v^{*} \in [0.0, 3.0]$, running forward
\item Cheetah-Dir: move forward and backward
\item Ant-Dir: move forward and backward
\item Ant-Goal: navigate to a target goal position on the 2D grid with $r \in [0.0, 3.0], \theta \in [0^{\circ}, 360^{\circ}]$. $r$ and $\theta$ represent the radius and angle of the goal from the origin, respectively.
\item Walker-Rand-Params: the agent is initialized with some system dynamics parameters randomized and must move forward.
\item Humanoid-Dir: run a target direction $\theta \in [0^{\circ}, 360^{\circ}]$ on the 2D grid. \par
\end{itemize}

\subsubsection{MuJoCo parametric out-of-distribution environments} \par
\label{sec:mujoco-ood-details}

\begin{table}[t]
  \centering
  \caption{Out-Of-Distribution(OOD) MuJoCo task settings}
    \scalebox{0.9}{
    \begin{tabular}{cccc}
    \toprule
          & \textbf{Cheetah-Vel-OOD} & \multicolumn{2}{c}{\textbf{Ant-Goal-OOD}} \\
          & $v$     & $r$     & $\theta$ \\
    \midrule
    $ \mathcal{M}_{train}$ & $[2.0, 4.0]$ & $[1.0, 3.0]$ & $[0^{\circ}, 360^{\circ})$ \\
    $\mathcal{M}_{test}$ & $[1.0, 2.0], [4.0, 5.0]$ & $[0, 1.0] , [3.0, 4.0]$ & $[0^{\circ}, 360^{\circ})$ \\
    \bottomrule
    \end{tabular}%
    }
  \label{tab:ood-tasks}%
\end{table}%

\begin{itemize}
\item Cheetah-Vel-OOD: achieve a target velocity running forward, meta-training velocity $v_{train} \in [2.0, 4.0]$, meta-test velocity $v_{test} \in [1.0, 2.0] \cup [4.0, 5.0]$.
\item Ant-Goal-OOD: navigate to a target goal position on the 2D grid with $r_{train} \in [1.0, 3.0]$, $r_{test} \in [0.0, 1.0] \cup [3.0, 4.0]$, $\theta \in [0^{\circ}, 360^{\circ}]$. $r$ and $\theta$ represent the radius and angle of the goal from the origin, respectively. \par
\end{itemize}

\subsubsection{MuJoCo non-parametric environments} \par
\label{sec:mujoco-non-parametric-details}

\begin{figure}[t]
  \centering
    \subfloat[\textit{Cheetah-Multi-Task}: Velocity, Goal, Stand, Jump] {\includegraphics[width=0.48\textwidth]{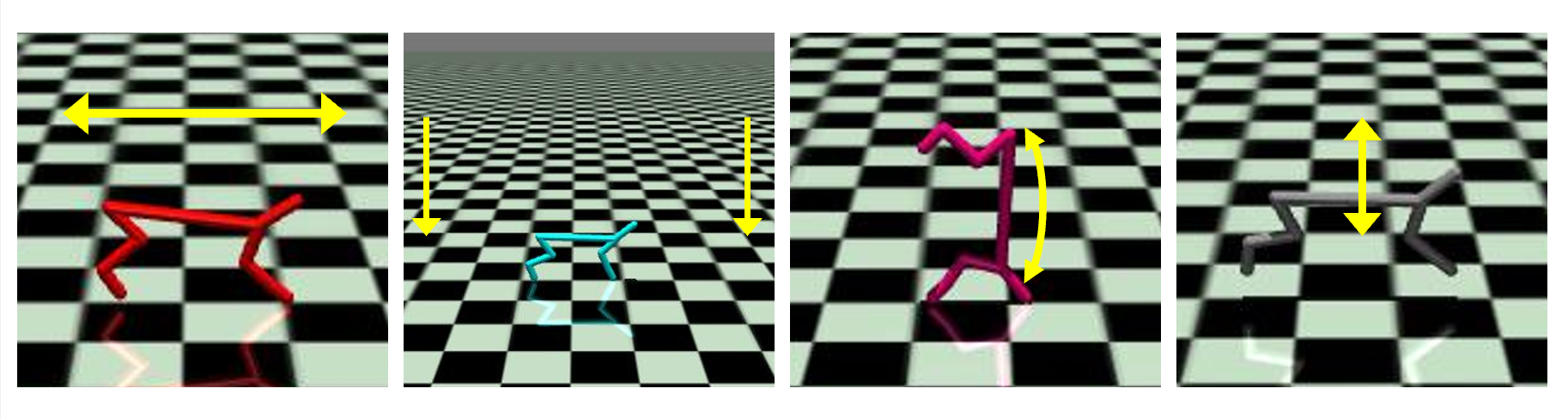}}
    \\
    \subfloat[\textit{Ant-Multi-Task}: Velocity, Goal, Jump] {\includegraphics[width=0.36\textwidth]{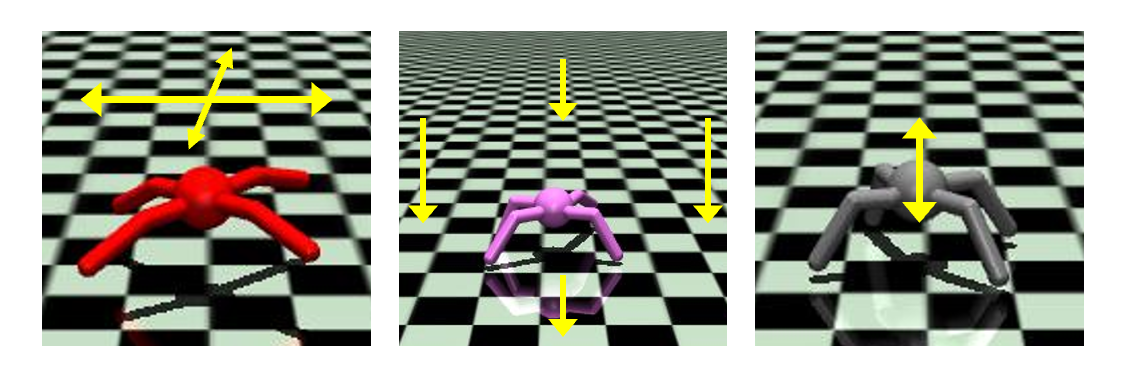}}
  \caption{Illustration of non-parametric benchmarks}
	\label{fig:mujoco-task-illustration}
\end{figure}

\begin{itemize}
\item Cheetah-Multi-Task: the agent needs to solve 4 continuous control tasks that are qualitatively distinct: Cheetah-Velocity, Cheetah-Goal, Cheetah-Stand and Cheetah-Jump. Details see Table \ref{tab:cheetah-multi-task-description}.

\item Ant-Multi-Task: the agent needs to solve 3 continuous control tasks that are qualitatively distinct: Ant-Velocity, Ant-Goal and Ant-Jump. Details see Table \ref{tab:ant-multi-task-description}.
\end{itemize}


\begin{itemize}
\item Reach: the robot arm should reach different initially unknown 3D goal locations.
\item Push: the robot arm should push objects to initially unknown 3D goal locations.
\item Pick-and-Place: the robot arm should pick up an object and place it near a target goal location in 3D space.
\end{itemize}

\subsubsection{Hyperparameters}
\label{sec:hyperparameters}
We implement our algorithm based on the PyTorch framework \cite{pytorch} and train the model on Tesla V100 GPUs. The hyperparameters on MuJoCo \emph{Cheetah-Vel}, \emph{Cheetah-Multi-Task} and Meta-World ML1 \emph{Push} are given in the table below. Please refer to the source code for more details.

\begin{table*}[t]
\label{tab:hyperparameters}
  \centering
  \caption{MoSS hyperparameters}
    \begin{tabular}{lccc}
    \toprule
          & \multicolumn{1}{l}{Cheetah-Vel} & \multicolumn{1}{l}{Cheetah-Multi-Task} & \multicolumn{1}{l}{Push}\\
    \midrule
       \textbf{General} &       &       &  \\
    \midrule
    n\_train\_tasks & 100   & 70    & 50 \\
    n\_eval\_tasks & 30    & 35    & 50 \\
    use\_normalized\_env & True     & True     & True \\
    num\_train\_epochs & 501   & 3001  & 2501 \\
    num\_train\_tasks\_per\_episode & 30    & 70    & 30 \\
    batch\_size\_reconstruction & 256   & 256   & 512 \\
    batch\_size\_policy & 256   & 256   & 256 \\
    max\_path\_length & 200   & 200   & 200 \\
    num\_eval\_trajectories & 1     & 1     & 1 \\
    \midrule
    \textbf{Policy} &       &       &  \\
    \midrule
    sac\_layer\_size & 300   & 300   & 300 \\
    policy\_net\_lr & 3e-4 & 3e-4 & 3e-4 \\
    automatic\_entropy\_tuning & False    & False    & False\\
    sac\_alpha & 0.2   & 0.2   & 0.2 \\
    reward\_scale & 1     & 1     & 10 \\
    latent\_size & 5     & 8     & 5 \\
    num\_training\_steps\_policy & 2048  & 2048  & 2048 \\
    num\_transitions\_initial & 200   & 200   & 800 \\
    num\_transitions\_per\_episode & 200   & 200   & 800 \\
    time\_steps & 64    & 64    & 64 \\
    use\_trajectory\_sample\_sac & True    & False    & True\\
    retain\_hidden & -   & -   & True\\
    use\_reward\_normalization & True    & True    & True\\
    bayes\_adaptive & True    & True    & True\\
    policy\_dim\_preprocess & True    & True    & False\\
    state\_embed\_dim & 32    & 32    & -\\
    latent\_embed\_dim & 32    & 32    & -\\
    clip\_grad\_policy & True    & True    & False\\
    max\_grad\_norm\_policy & 0.5   & 0.5   & -\\
    \midrule
    \textbf{VAE} &       &       &  \\
    \midrule
    lr\_encoder & 3e-4 & 3e-4 & 3e-4 \\
    lr\_decoder & 3e-4 & 3e-4 & 3e-4 \\
    num\_classes & 1     & 4     & 1 \\
    num\_training\_steps\_reconstruction & 128   & 128   & 128 \\
    clip\_grad\_vae & True    & True    & False\\
    max\_grad\_norm\_vae & 1.0     & 1.0     & -\\
    use\_global\_prior & False    & False    & False\\
    gmvae\_training\_complete\_path & False    & False    & True\\
    \bottomrule
    \end{tabular}%
  \label{tab:ood-tasks}%
\end{table*}%

\end{document}